\documentclass[10pt,twocolumn,letterpaper]{article}
\PassOptionsToPackage{dvipsnames,table}{xcolor}

\usepackage[pagenumbers]{iccv} 

\definecolor{iccvblue}{rgb}{0.21,0.49,0.74}
\usepackage[pagebackref,breaklinks,colorlinks,allcolors=iccvblue]{hyperref}

\def\paperID{*****} 
\def\confName{ICCV}
\def\confYear{2025}

\title{Generating Synthetic Data via Augmentations for Improved Facial Resemblance in DreamBooth and InstantID}

\author{Koray Ulusan\\
University of Tuebingen\\
{\tt\small koray.ulusan@uni-tuebingen.de}
\and
\and
Benjamin Kiefer\\
LOOKOUT\\
University of Tuebingen\\
{\tt\small benjamin@lookout.team}
}

\usepackage{gensymb}
\usepackage{caption}

\usepackage{dblfloatfix}

\usepackage{caption}
\usepackage{cuted}
\usepackage{xcolor}

\usepackage{booktabs}
\usepackage{multirow}

\newcommand{\colorboxinline}[1]{\textcolor[HTML]{#1}{\rule{1.2ex}{1.2ex}}}

\begin{document}


\maketitle

\begin{strip}
  \centering
  \includegraphics[width=0.91\linewidth]{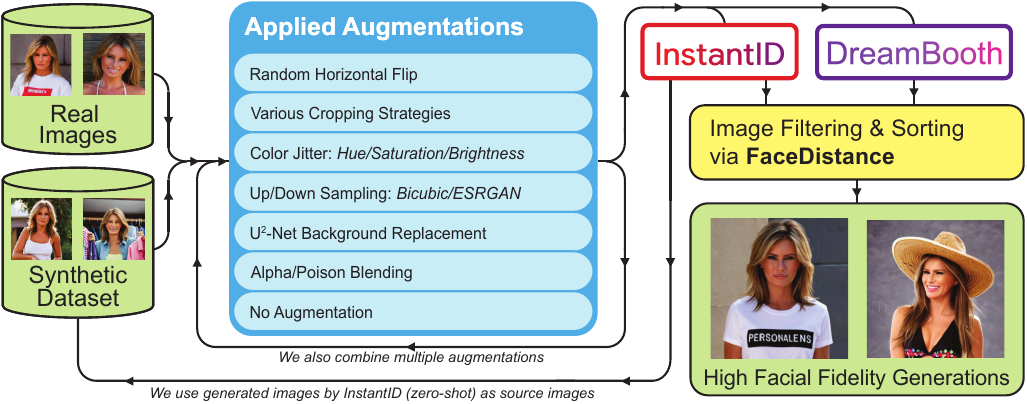}
  \captionof{figure}{%
  Pipeline for generating personalized portraits using synthetic images enhanced through classical and generative augmentations to improve identity resemblance in DreamBooth and InstantID outputs.
  }
  \label{fig:generalpipeline}
\end{strip}


\begin{abstract}%
%
Personalizing Stable Diffusion for professional portrait generation from amateur photos faces challenges in maintaining facial resemblance. This paper evaluates the impact of augmentation strategies on two personalization methods: \textnormal{\mbox{DreamBooth}} and \textnormal{\mbox{InstantID}}. We compare classical augmentations (flipping, cropping, color adjustments) with generative augmentation using InstantID’s synthetic images to enrich training data. Using \textnormal{\mbox{SDXL}} and a new 
\textnormal{\mbox{FaceDistance}} metric based on FaceNet, we quantitatively assess facial similarity. Results show classical augmentations can cause artifacts harming identity retention, while InstantID improves fidelity when balanced with real images to avoid overfitting. A user study with 97 participants confirms high photorealism and preferences for InstantID’s polished look versus DreamBooth’s identity accuracy. Our findings inform effective augmentation strategies for personalized text-to-image generation.%
\\
\end{abstract}

\vspace{-3.7em}


\section{Introduction}
Personalized text-to-image generation has gained traction with the rise of models like Stable Diffusion (SD). However, training SD on small, user-specific datasets presents challenges, such as identity retention, overfitting, and artifact generation. Augmentation techniques are widely used in deep learning to improve generalization, but their role in personalized text-to-image generation is underexplored.

In this work, we analyze how augmentations affect personalized SD models trained via few-shot and zero-shot methods, focusing on DreamBooth and InstantID. We examine their impact on model performance and whether they improve the realism and consistency of generated images.

In particular, we analyze both classical and generative augmentation strategies to bridge the gap between limited real data and high-fidelity synthetic outputs. By refining facial features and preserving identity through targeted GenAI-based augmentations, such as InstantID, we aim to improve the applicability of personalized generation in scenarios where synthetic data must closely mirror real-world characteristics.
We analyze under which conditions we can ensure that ``GenAI outputs improve GenAI outputs'', avoiding a data quality collapse, providing best practices and heuristics.

Our contributions include:
\begin{itemize}
    \item Analysis of classical augmentation techniques such as flipping, cropping, color enhancement, and background modifications.
    \item Using InstantID as a fast way of enhancing the user-specific dataset using the diffusion model itself. 
    \item We conduct a survey to evaluate how white-collar workers perceive personalized generations from DreamBooth and InstantID under various augmentation strategies.
\end{itemize}

\section{Background and Related Work}

In this section, we present the foundational concepts and prior research relevant to our work on augmentation techniques for few-shot personalization in diffusion models.

\subsection{Text-to-Image Diffusion Models}

Text-to-image diffusion models generate high-quality images from natural language descriptions by gradually denoising random Gaussian noise guided by text embeddings \cite{Rombach2022, Nichol2021Feb}. Stable Diffusion \cite{Rombach2022} employs a latent diffusion approach that operates in a compressed latent space rather than pixel space, reducing computational requirements while maintaining generative quality.

\subsection{Subject-Driven Image Generation}

Subject-driven image generation creates images featuring specific subjects with high fidelity while maintaining their identity across contexts \cite{Ruiz2022Aug}. Key approaches include:

\textbf{DreamBooth} \cite{Ruiz2022Aug} fine-tunes the U-Net of Stable Diffusion using 3-5 images of a specific subject. It preserves the semantic prior through class-specific prior preservation loss and uses a rare token with weak prior to refer to the subject with the prompt format ``a [$V$] [class noun]''.


\textbf{InstantID} \cite{Wang2024Jan} is a zero-shot method combining facial feature extraction and text conditioning. It uses 5 facial landmarks to control the face’s position and orientation, providing greater control over the output.

We standardize experiments by using the same SDXL model for both methods to ensure a fair comparison.

\subsection{Image Augmentation Techniques}

\subsubsection{Classical Image Augmentations}

Classical image augmentations include geometric transformations (flipping, rotation, scaling, cropping), photometric adjustments (brightness, contrast, saturation, hue), and noise injections (Gaussian, salt-and-pepper). These predefined techniques preserve semantic integrity while adding controlled diversity to expand limited training data.

\subsubsection{Augmentations in Diffusion Models}

Data augmentation enhances diffusion model performance while reducing computational demands \cite{Trabucco2023Feb}. Key approaches include mixing-based augmentations that interpolate between existing samples \cite{Islam2024Apr} and consistency regularization techniques that enforce invariance to specific transformations \cite{Islam2025Mar, Liew2022Oct}. Our work investigates these techniques specifically for few-shot personalization applications.

\subsection{Face Processing Approaches}

FaceNet \cite{Schroff2015Mar} maps facial images to a 128-dimensional embedding space, where similar faces are positioned closely together. The standard pipeline uses MTCNN \cite{Zhang2016Apr} for face detection before embedding generation, with cosine distance metrics for similarity assessment \cite{Serengil2024Mar}.

Alternative approaches include faceswapping methods \cite{Li2019Dec, Agarwal2022Aug} and augmented reality techniques for virtual try-on applications \cite{GlamTry2024}. While these provide real-time capabilities, they often lack the flexibility and integration capabilities of diffusion-based approaches.

Our research builds on these foundations to investigate how strategic data augmentation can improve few-shot personalization in diffusion models, focusing on identity preservation and recontextualization.


\begin{figure*}[t]
    \centering
    \begin{minipage}{0.191\textwidth}
        \centering
        \includegraphics[width=\linewidth]{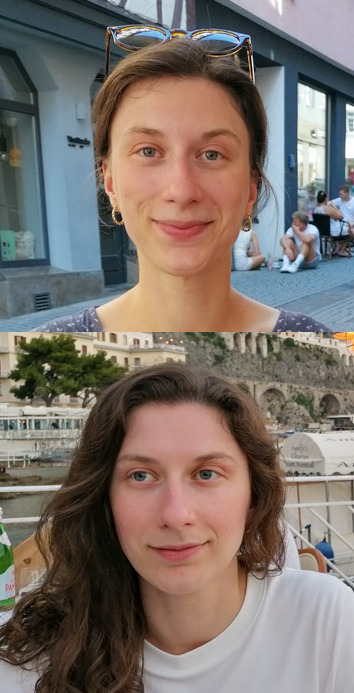}
        \subcaption{Real Images\\\phantom{.}}
    \end{minipage}
    \hfill
    \begin{minipage}{0.373\textwidth}
        \centering
        \includegraphics[width=\linewidth]{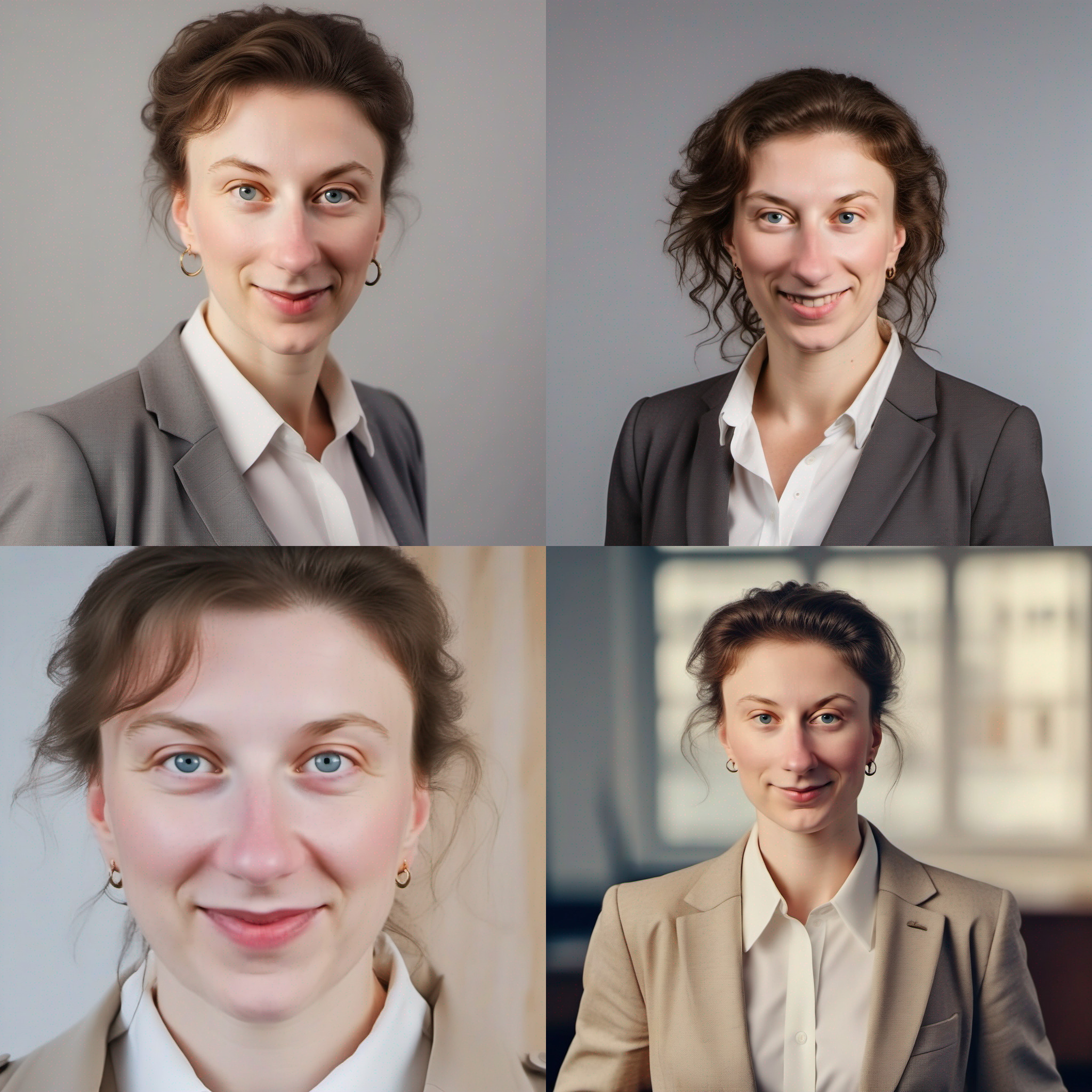}
        \subcaption{DreamBooth results with classical augmentations (crop, resize, and color)}
    \end{minipage}
    \hfill
    \begin{minipage}{0.373\textwidth}
        \centering
        \includegraphics[width=\linewidth]{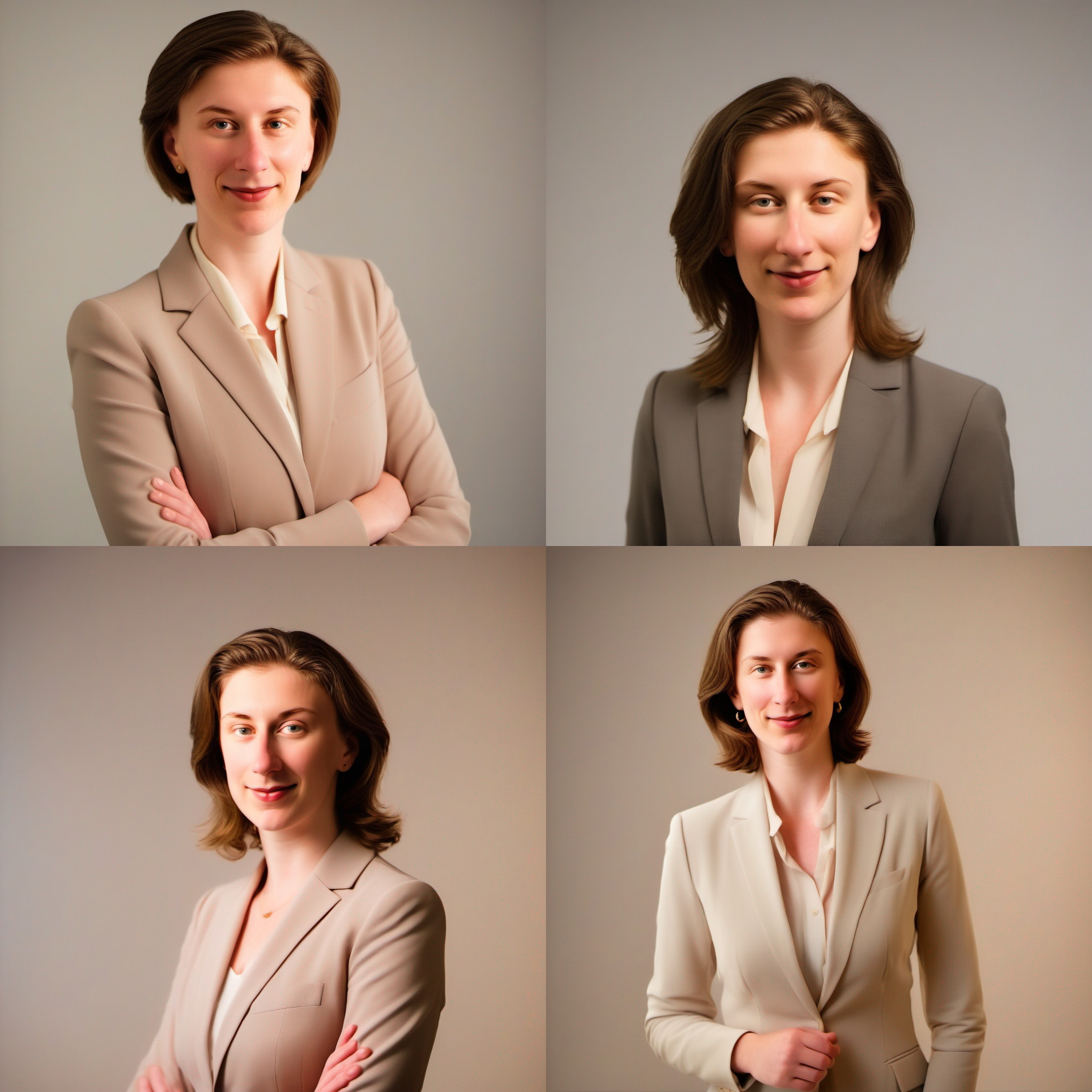}
        \subcaption{DreamBooth results with GenAI augmentations \& without classical augmentations}
    \end{minipage}
    \caption{Example Improvement: DreamBooth Results Comparing Classical vs. Generative Augmentations for Subject Dataset \textit{Vacation-Anna}.
    The model is prompted with \textit{default prompt} with batchsize 4. The results are \textbf{not} cherry-picked to resemble the downstream application use. Although (b) is visually more interesting, the method in (c) is more consistent across many subject datasets.}
    \label{fig:genai-results}
\end{figure*}

\section{Methodology}

We use augmentations across various Subject Datasets to see if there is an overall improvement in generated pictures.

\subsection{Subject Datasets}

Our internal dataset includes 3 to 15 images per participant, with 10 participants. To maintain naturalistic data collection, we instructed them: \textit{``Can you send me portrait/selfie-style photos of your face in different places? The more different places, the better.''} By avoiding strict guidelines, we ensured the images reflect realistic user behavior. Consequently, our findings align well with real-world data distributions, improving transferability and applicability. The names used to refer to subjects in this dataset are \emph{pseudonyms} and do not correspond to real individuals.

Our internal dataset exhibits diverse environmental conditions, facial orientations, and image qualities, ensuring variability that reflects real-world scenarios. The images include varied backgrounds, lighting, and subject behaviors, making the dataset both representative and robust. For example, some images feature cluttered or irregular backgrounds (e.g., \textit{Baker-Zoe}, \textit{Bottle-Hugo}), while others are from controlled settings (\textit{Biometric-Kora}). Gaze variation is present: \textit{Doctor-Nina} looks away from the camera, and \textit{3D-Gary} shows dynamic head movements from video. Differences in appearance and accessories also appear, such as \textit{Farmer-Lisa} wearing a helmet and \textit{Staircase-Judy} wearing makeup. Lighting varies from well-lit (\textit{Vacation-Anna}) to suboptimal (\textit{2024-Kora}), enhancing realism. These factors make our dataset valuable for evaluating model performance in unconstrained, real-world conditions.


\subsection{Dataset Augmentations}
\label{subsec:dataset_aug}
We apply augmentations individually to evaluate each technique's performance improvement independently.

\textbf{Classical Augmentations\quad} Standard techniques include: (i) \textit{Random Horizontal Flip} with $p\in\{0, \frac{1}{2}, 1\}$, and (ii) \textit{Color Jitter} varying brightness, saturation $(\pm 5, \pm 15)$ and hue $(\pm 5\degree, \pm 15\degree)$.

\textbf{Background Augmentation\quad} We use U$^2$-Net \cite{Qin2020May} for subject isolation, testing both base and human segmentation models. Backgrounds include flat colors, Wikimedia Patterns \cite{wikiiiiii}, and Studio Backdrops (handpicked simple pictures from BG-20k \cite{Li2020Oct}, combined with our internal dataset).

\textbf{Blending Techniques\quad} We separately evaluate \textit{Alpha Blending} and \textit{Poison Blending} through both automated and manual techniques.

\textbf{Resizing Methods\quad} We compare: (i) downsampling then upsampling, (ii) upsampling only, and (iii) original dimensions. Methods include ESRGAN \cite{Wang2018Sep}, Lanczos, and Bilinear.

\textbf{Cropping Strategies\quad} Five approaches: (i) SDXL dimensions \cite{Podell2023Jul}, (ii) automated center cropping to 1MP, (iii) downsample-then-crop to 1MP at various aspect ratios, (iv) manual eight-variation cropping, and (v) MTCNN face-based cropping.

\textbf{Color Adjustment\quad} Adobe Lightroom auto-adjustment enhances visual quality.

\textbf{Generative Augmentation\quad} Using InstantID, we generate new subject images with prompts from \texttt{dolphin 2.2.1 - Mistral 7B} \cite{BibEntry2025Mar234} and varied facial landmarks. Based on one or more input images of a person, we run them through the InstantID pipeline with augmented landmarks and LLM generated prompts. The landmarks are extracted from the input images and then slightly perturbed to preserve resemblance while introducing diversity. (Omitting this step results in limited expression diversity, with generated images often replicating the same facial expression—an effect clearly visible in the rightmost examples of Figure~\ref{fig:2-step-generation}.) This process yields a synthetic dataset of personalized subject images, which is subsequently used for downstream DreamBooth training to enhance model specificity and subject fidelity.

\subsection{Hardware, Software, and Hyperparameters}

All experiments were conducted on a single NVIDIA GeForce RTX 3090 with 24GB VRAM.  We use \texttt{sd-scripts}\cite{BibEntry2025Mar3} for DreamBooth and \texttt{ComfyUI\_InstantID}\cite{BibEntry2025Ma54} for InstantID experiments, inheriting all bias in their pipeline, if any. 

Results of DreamBooth finetuning a diffusion model (DM) greatly depends on the DMs ability of generating images. We use \texttt{RealVisXL\_V4.0} \cite{BibEntry2025Mar2}, which is a community finetune of SDXL for realistic image generation.



The \textbf{default prompt} is ``A professional headshot of a \textit{subject} wearing a suit in a well-lit studio, DSLR.'' Empirically, including gender as \textit{man} or \textit{woman} imparts western cultural gender traits, which we prefer.

For DreamBooth, we use ``a [$V$] [man$\mid$woman]'', where [$V$] is the rare token. InstantID uses no special prompt and works with any text. LLM-generated prompts benefit both.

\subsection{FaceDistance Metric}

To quantify facial similarity in generated images, we employ the \textbf{FaceDistance} metric based on FaceNet embeddings \cite{Schroff2015Mar}. FaceNet projects facial images into a 128-dimensional hyperspherical embedding space where spatial proximity reflects facial similarity.

Given a dataset of subject images, we first compute their FaceNet embeddings and derive the \textit{mean face vector}, denoted as $\bar{v}_{\text{real}}$. For any generated image, we then calculate the cosine distance between its embedding vector and $\bar{v}_{\text{real}}$. A lower cosine distance implies greater resemblance to the real subject.
FaceDistance serves as a practical tool for:
\begin{itemize}
  \item Coarsely ranking generated images by similarity (lower distances are better),
  \item Discarding the top $k$\% of distant embeddings to improve personalization quality\footnote{Empirical evidence supports $k = 15$ for datasets with $n \geq 8$; independently, values above the 80th percentile $(q_{0.80})$ may be discarded.},
  \item Identifying failure cases such as off-subject generations or severe artifacts.
\end{itemize}

In our experiments, the cosine distance between real images and $\bar{v}_{\text{real}}$ (within a person's dataset) has a mean of $0.13$, a maximum of $0.22$, and a minimum of $0.05$.

For a detailed discussion of how FaceDistance enhances downstream applications such as large-scale image filtering and user satisfaction, see Suppl.~\ref{app:facedistance-downstream-apps}.

\section{Experiment Results}

We try to achieve higher facial similarity via  DreamBooth and InstantID using highlighted augmentations.


Despite using a realistic image generation model, achieving photorealistic face generation remains challenging without strict constraints on subject images. We relaxed many constraints to improve usability, since expecting average users to compile datasets without understanding image generation is difficult. High subject fidelity is crucial for effectiveness in downstream tasks, as humans are more sensitive to facial feature variations than to textures.


A major issue with unconstrained datasets is that the images may poorly represent the person. This is similar to struggling to recognize someone in real life based only on a few photographs. We observe this with small datasets where $\operatorname{size} \leq 3$. In such cases, generated images reflect the dataset well—outsiders unfamiliar with the person often consider the images accurate. However, those who know the person find the generated images to be poor representations.

\subsection{DreamBooth}

We configure our hyperparameters to prioritize high facial fidelity over recontextualization capabilities. Identity preservation is challenging in DreamBooth, so we prefer to overfit for strong subject fidelity, accepting limited generation diversity.


The common theme in augmentations is that if the augmented image has any kind of artifact/anomaly, then the rare token will be associated with it. The supporting observations are 
(i) When background is replaced with a geometric pattern (from wikimedia patterns \cite{wikimedia_patterns}), the model will focus on learning the pattern than the subject 
(ii) When image is upscaled with ESRGAN, the texture ESRGAN introduces say present in generations 
(iii) the masks generated with U$^2$-Net is not pixel-perfect. and results in a mix around hair/air boundary. This mix becomes associated with the subject. 
The human segmentation models training data was not highly accurate around hairs but was better in identifying body parts. The base model is performs better around hair and was overall better. The robustness of human segmentation model is not needed. 
(iii) any kind of \textit{color jitter} is visible in generated images. For example the saturation change of $0.1$ is clearly present in generations. 
(iv) using Adobe Lightroom as a preprocessing step resulted in better color graded generations compared to non-preprocessed datasets. 
(v) datasets with low contrast (e.g. exclusively Polaroid pictures) resulted in copying the photography style/lighting from the pictures — though this can be attributed to our hyperparameter configuration. 

Because of the low recontextualization capabilities, backgrounds becomes highly associated with the rare token. Replacing the background with \textbf{Pastel Colors}
[\colorboxinline{55efc4}
\colorboxinline{81ecec}
\colorboxinline{74b9ff}
\colorboxinline{a29bfe}
\colorboxinline{ffeaa7}
\colorboxinline{fab1a0}
\colorboxinline{ff7675}
\colorboxinline{fd79a8}]
and \textbf{Rainbow Colors} 
[
\colorboxinline{2ecc71}
\colorboxinline{3498db}
\colorboxinline{9b59b6}
\colorboxinline{f1c40f}
\colorboxinline{e67e22}
\colorboxinline{e74c3c}]
led to eccentric and often unrealistic images, with the latter occasionally generating pictures without subjects. \textbf{Gray} (range [0--255]) offered the highest resemblance to the subject, while \textbf{Dark Gray} (range [0--127]) caused the model to disassociate the subject from its context. Because of problems with U$^2$-Net, \textbf{Light Gray} (range [128--255]) background outperformed Dark Gray, especially in bright environments. We believe replacing all backgrounds with similar themed (colorful/gray) background degraded models recontextualisation capability. \textbf{Wikimedia Patterns} slowed down learning and degraded the image quality across all generations. Lastly, \textbf{Studio Backdrops} introduced irregularities that reduced the quality of the generated images which can be thought as similar to Wikimedia Patterns because backdrops has patterns. The regular patterns and backdrops captured models focus more than the subject, which is undesirable when the goal is to teach the subjects face.

\begin{figure}[htb]
    \centering

    \begin{subfigure}[b]{0.32\linewidth}
        \includegraphics[width=\linewidth]{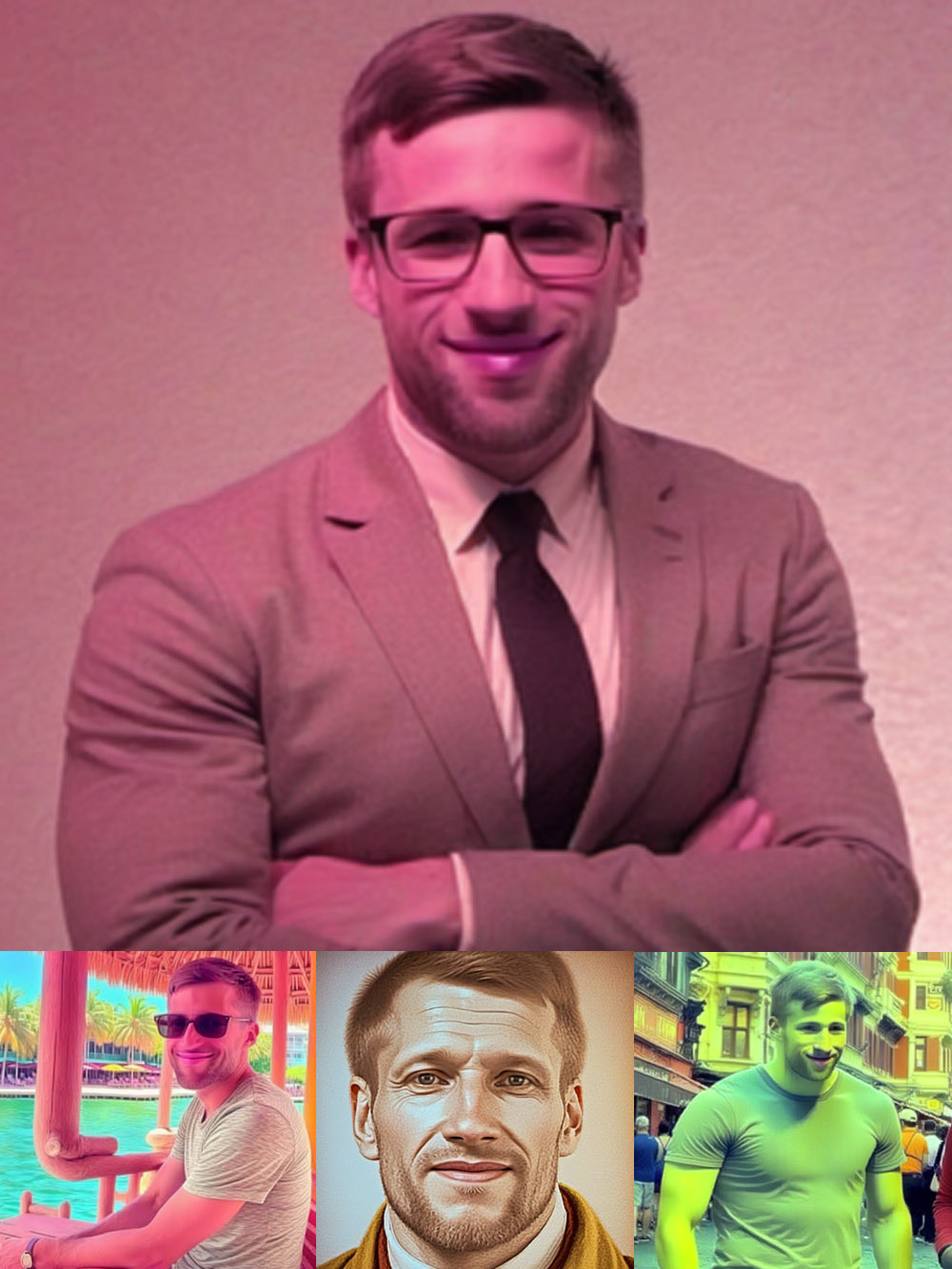}
        \caption{Hue Shift}
    \end{subfigure}
    \hfill
    \begin{subfigure}[b]{0.32\linewidth}
        \includegraphics[width=\linewidth]{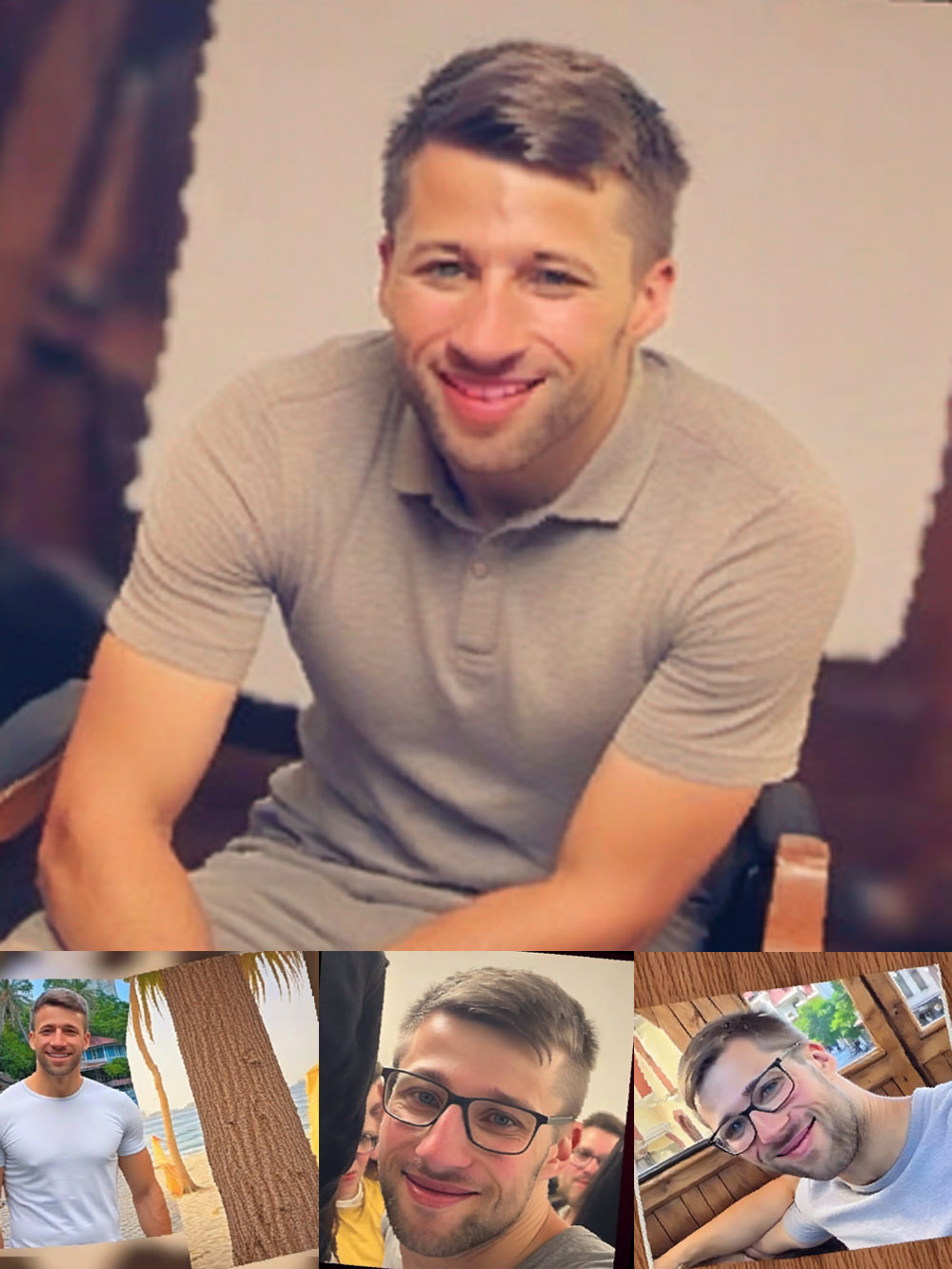}
        \caption{Random Rotation}
    \end{subfigure}
    \hfill
    \begin{subfigure}[b]{0.32\linewidth}
        \includegraphics[width=\linewidth]{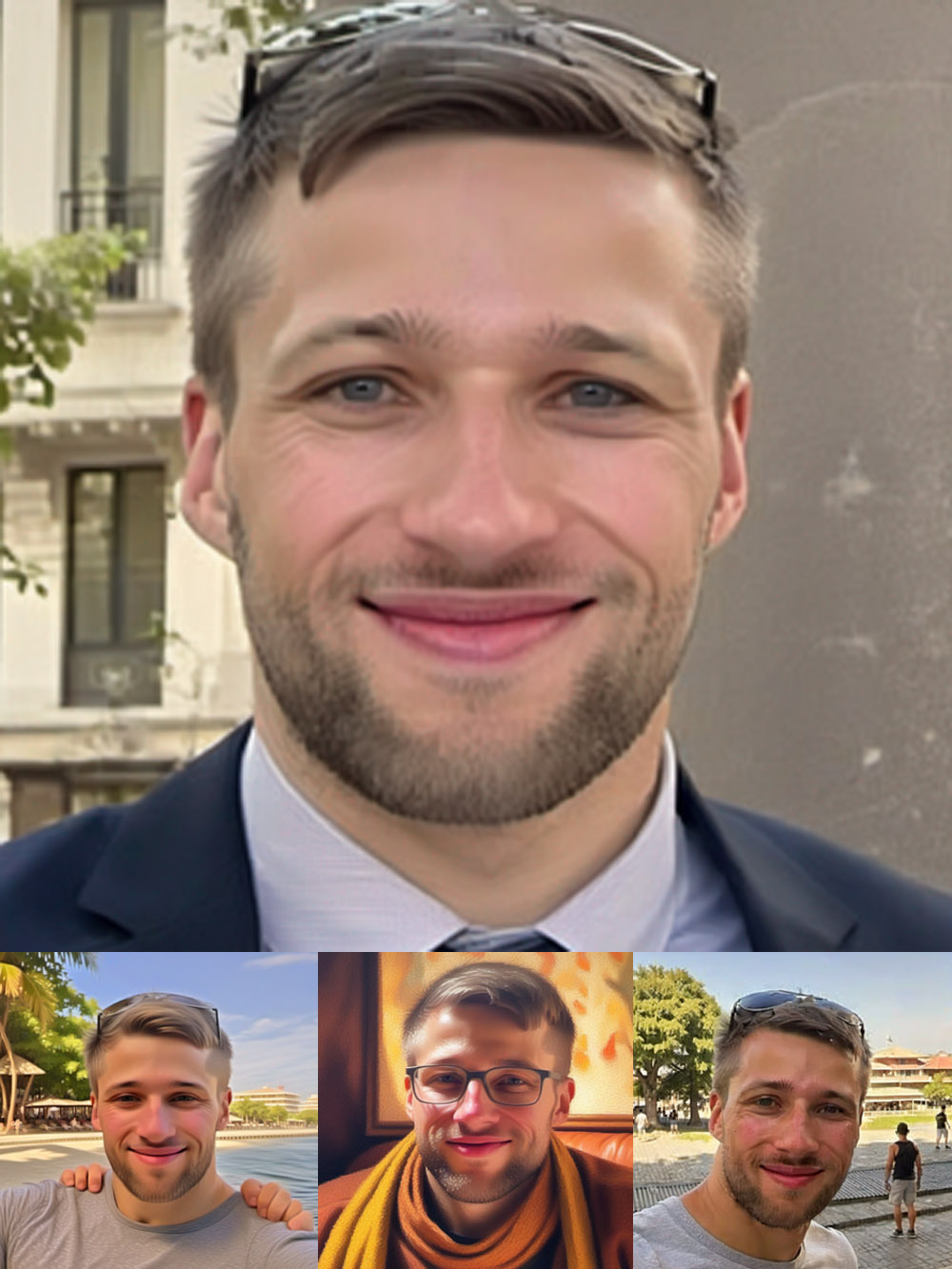}
        \caption{Saturation Jitter}
    \end{subfigure}
    
    \caption{Generated Artifacts from Augmentations. Color jitter (a, c) introduces unnatural hues and saturation; rotation (b) distorts alignment and adds padding objects.}
    \label{fig:rot_color_comparison}
\end{figure}

\textbf{Random Horizontal Flip} slowed learning due to face asymmetry, which confused facial features. \textbf{Random Rotation} introduced unnatural alignments and visual artifacts like objects or black padding bars. \textbf{Color Jitter} caused erratic generations, as brightness, contrast, saturation, and hue changes were linked to rare tokens. (Figure~\ref{fig:rot_color_comparison})

Both \textbf{Alpha Blending} and \textbf{Poison Blending} are discouraged, as they require careful manual processing to achieve good results. These techniques are not straightforward to apply and can lead to undesirable artifacts if mishandled.

\begin{figure}[htb]
    \centering
    \begin{subfigure}[b]{0.49\linewidth}
        \centering
        \includegraphics[width=\linewidth]{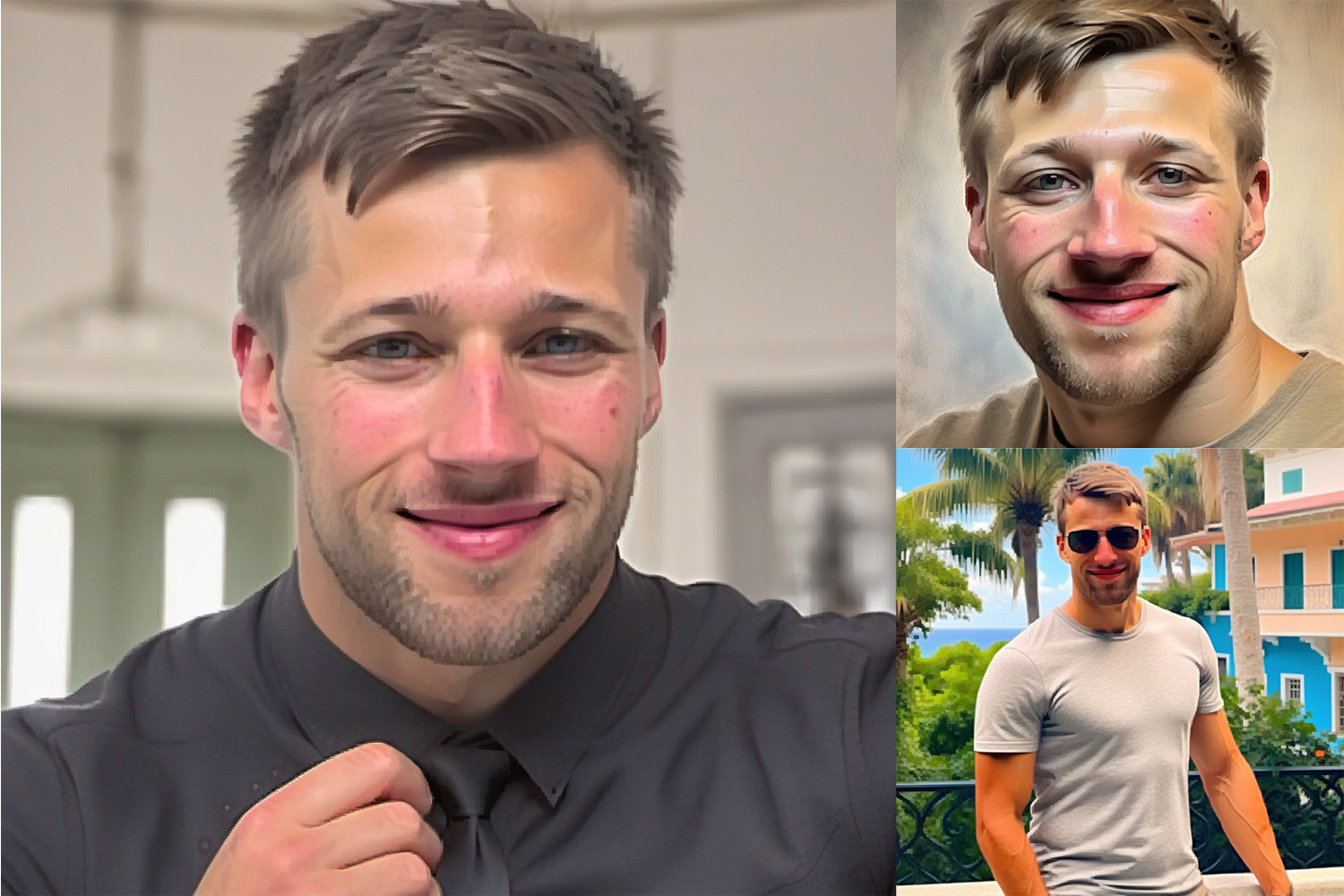}
        \caption{User Provided Dataset}
    \end{subfigure}
    \hfill
    \begin{subfigure}[b]{0.49\linewidth}
        \centering
        \includegraphics[width=\linewidth]{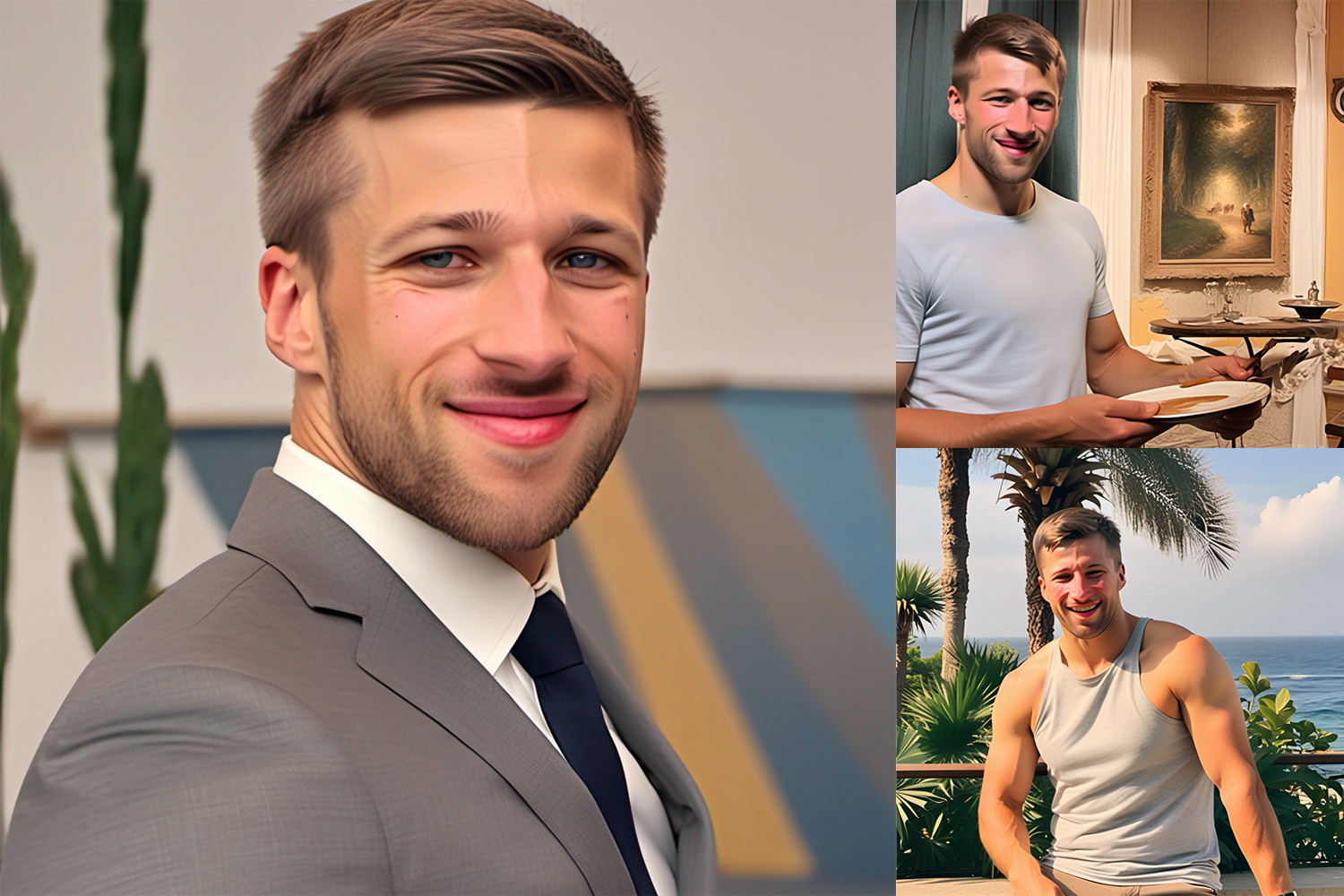}
        \caption{ESR-GAN Upscaled Dataset}
    \end{subfigure}
    
    \vspace{0.4em}
    
    \begin{subfigure}[b]{0.49\linewidth}
        \centering
        \includegraphics[width=\linewidth]{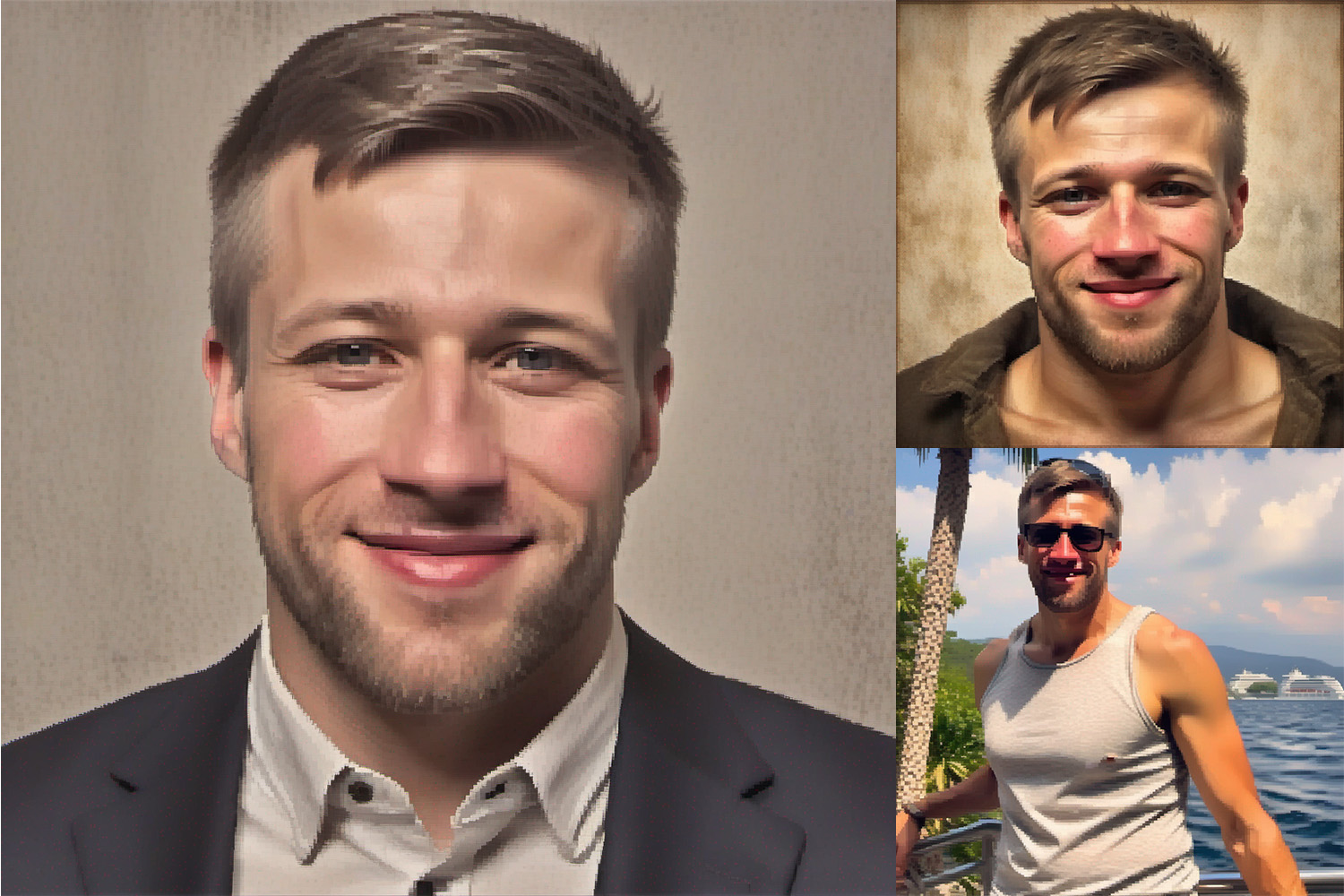}
        \caption{Downsampled Dataset}
    \end{subfigure}
    \hfill
    \begin{subfigure}[b]{0.49\linewidth}
        \centering
        \includegraphics[width=\linewidth]{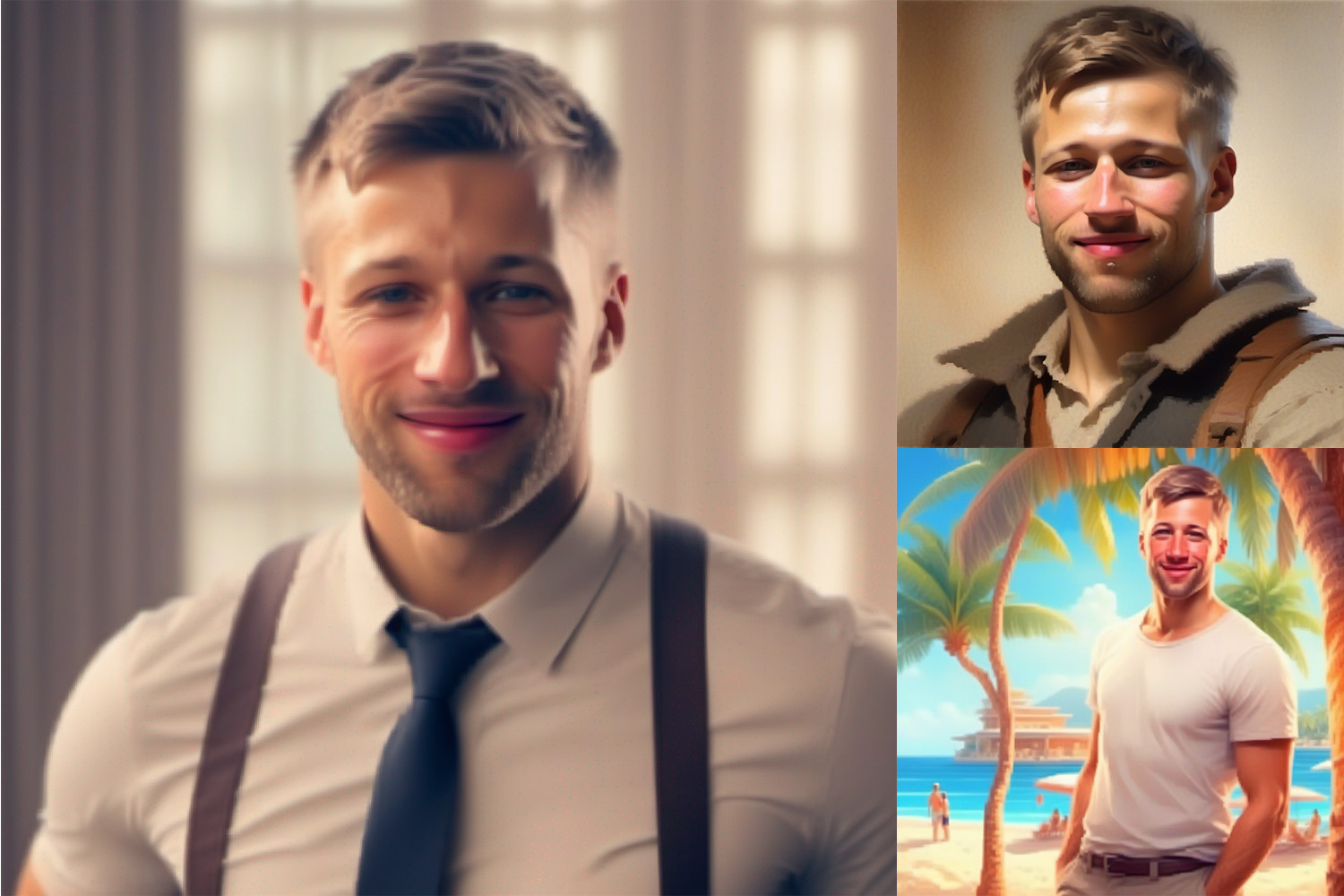}
        \caption{Down then Upsampled Dataset}
    \end{subfigure}
    
    \caption{Comparison of generated image quality when conditioned on subjects from datasets with different resolution manipulations.  
    The subfigure captions indicate the type of conditioning dataset used for generation, while the \textbf{images show the resulting generated samples.}
    (a) Original dataset images ($\approx$ 0.7MP). 
    (b) Upsampling artifacts, notably around the eyes. 
    (c) Simulated low-res input via 0.5$\times$ Lanczos downsampling; compare with (a). 
    (d) Downsampled (0.5$\times$) then upsampled (2$\times$) with bicubic interpolation—results show smoothed textures.
    }
    \label{fig:resolution_comparison}
\end{figure}

Images around \textbf{1 megapixel} (MP) performed best, providing a balanced resolution for high-quality generation. We believe this is due to SDXL’s training resolutions (see Appendix I of \cite{Podell2023Jul}) being  $\approx$ 1 MP, and the model tends to perform poorly outside this range where it lacks sufficient training data. \textbf{Upscaling with ESR-GAN} introduced visible artifacts, especially around facial features. \textbf{Upscaling with Lanczos} was effective, particularly when starting from larger images. However, if the initial dataset contained low-resolution images, the generated images exhibited facial blurring due to the nature of the Lanczos algorithm. The difference between bicubic and Lanczos was negligible. \textbf{Downscaling} resulted in more pixelated generations compared to using original-sized images. It should be noted that our testing output resolution was $1024\times 1024$. Figure~\ref{fig:resolution_comparison} demonstrates these findings.

\textbf{InstantID Augmentation}\quad  
Datasets augmented with InstantID yield clearly superior performance. The added images need to be diverse (i.e., generated with various text conditioning and different keypoint images). Since we trade recontextualization abilities for increased facial similarity, generating the same person in similar contexts is beneficial. DreamBooth achieves similar facial similarity compared to InstantID but allows for greater control. The rigidity caused by the keypoint images is eliminated. \textit{2-step generation} method is twice as computationally expensive than using InstantID once and is not practical as \textit{face replacement} when augmenting for Dreambooth finetuning usage. In such scenarios, automated \textit{face replacement} emerges as a more efficient and suitable alternative. Additionally, achieving proper prompt diversity can be challenging. We preferred InstantID over DreamBooth.

The ratio of real to InstantID-generated images depends entirely on the diversity of the generated images. No single concept should comprise more than $25\%$ of the dataset. For example, if images labeled as ``a [$V$] man in a library'' exceed $25\%$, DreamBooth training will associate the rare token with the concept. This results in a final DreamBooth model that is unusable due to a complete loss of recontextualization ability caused by overfitting.

Since InstantID generations are highly realistic, additional images can be generated to better represent the subject during DreamBooth training. Using the same diffusion model for both InstantID and DreamBooth helps integrate the subject effectively without changing its context, keeping the dataset distribution closer to the diffusion model’s generation space.

\subsubsection{FaceDistance} 

\begin{figure}[htb]
    \centering
    \includegraphics[width=0.98\linewidth]{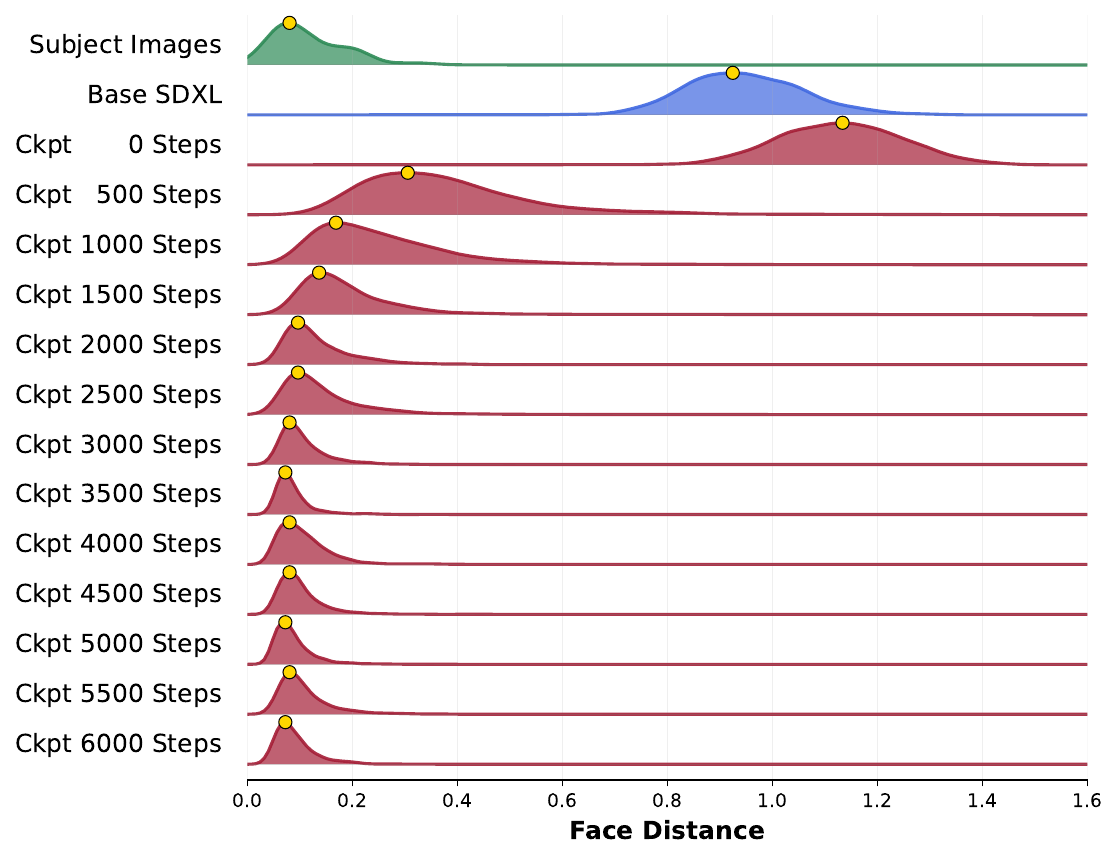}
    \caption{\textit{Face Distance Distributions Across Finetuning Steps for SDXL Real Using Closeup-Kora as Subject.}
    Kernel density estimates (KDEs) of face distances between 2{,}000 generated images and subject images are shown for different model checkpoints. The y-axis represents the image set generated by each model version, ordered by finetuning progress from ``\textcolor[HTML]{4F74E3}{Base SDXL}'' (blue, prompt: ``a man'') to finetuned checkpoints labeled ``\textcolor[HTML]{AB2E45}{Ckpt N Steps}'' (red, prompt: ``a [$V$] man'', where ``[$V$]'' denotes the rare token). The top row (green, ``\textcolor[HTML]{3C9261}{Subject Images}'') shows self-similarity of the subject images. FaceDistance ($x$-axis) measures identity similarity, with lower values indicating closer matches. The mode of each distribution is marked with a yellow dot.}

    \label{fig:baddbad}
\end{figure}

We selected the ``best'' DreamBooth checkpoint by generating images of ``a [$V$] man'' in different contexts for all checkpoints and ranking them using \textit{FaceDistance}. This discarded obviously bad checkpoints (e.g. anomalies in generations, unable to generate the subject, divergence) but is not able to rank ``good enough'' checkpoints within themselves. (Figure~\ref{fig:baddbad}) The FaceNet manifold isn't sensitive to very similar looking people. For a given hyperparameter configuration, a few tests show when the model will be converged to its best state (usually between 3k and 6k steps) and since FaceDistance isn't able to differentiate between them, FaceDistance isn't a useful tool for this purpose. 

Despite these challenges, FaceDistance appears to be functioning for loosely ranking generated images. This improves the user experience.

\subsection{InstantID} 
The effectiveness of InstantID is highly dependent on the quality and characteristics of the provided reference images.

\subsubsection{Face Embedding}
We conducted experiments to determine the optimal number of reference images that balances usability and facial similarity. Our findings confirm those of \cite{Wang2024Jan}, demonstrating that using multiple reference images results in increased facial similarity. When only one reference image is provided, the generated face is heavily influenced by the specific appearance captured in that single image. We attribute this limitation to insufficient information being extracted by the Face Encoder from a single perspective. Our analysis indicates that four reference images provide satisfactory results in most cases, with diminishing returns observed beyond this number. Since reference images are cropped and aligned before being processed by the face encoder, users have considerable flexibility in selecting images without compromising model performance.

\subsubsection{Landmarks Image}
\label{sec:landmarks-img}

We observe that facial landmarks exert strong conditioning influence, often rendering text prompts ineffective for controlling the subject's position. The generated image consistently replicates the face placement, orientation, and size specified by the provided keypoints, due to the five-point landmark system employed.

For practical use, users often struggle to understand how face positioning in the landmark image maps to the output. This disconnect often leads to dissatisfaction, even though the issue stems from suboptimal conditioning input.

\begin{figure*}
    \centering
    \includegraphics[width=0.98\linewidth]{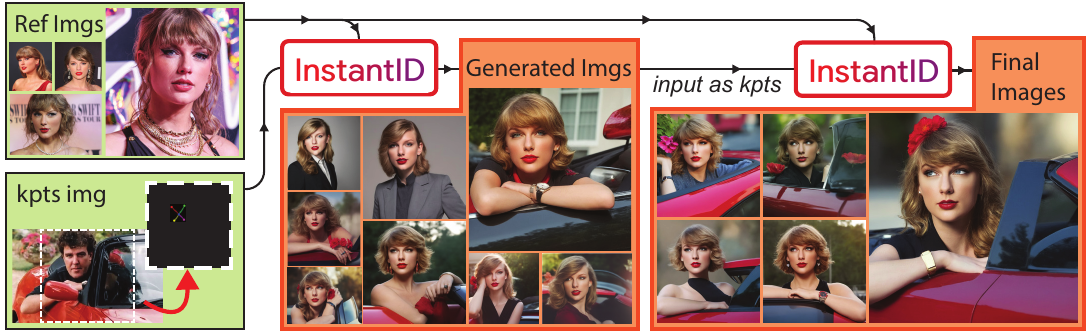}
    \caption{
     Two-Step Generation Pipeline. Initial outputs use a keypoints image ($s_{kpts}$) from another identity, often reducing facial similarity. Replacing $s_{kpts}$ with a prior output of the subject improves identity preservation while retaining pose. Using four reference images offers a good trade-off, as demonstrated in Appendix~\ref{app:facedistance-img-count}. Despite the ease-of-use in downstream applications, this limitation motivates our \textit{face replacement} method (Section~\ref{sec:landmarks-img}) for greater control.
    }
    \label{fig:2-step-generation}
\end{figure*}

To address this limitation, we propose two solutions: \textit{2-step generation} and \textit{face replacement}.

In \textbf{2-step generation} (Figure~\ref{fig:2-step-generation}), we collect \textbf{s}ubject reference images $(s_1, \dots, s_n)$ and a separate image representing the desired pose and composition $s_{kpts}$. These are used as reference images and the keypoints image, respectively. While the resulting output $s_{out}$ is generally satisfactory, using facial landmarks from one person to generate another reduces facial similarity due to structural differences in the five keypoints (eyes, nose, mouth). We hypothesize this stems from imbalanced conditioning weights. Performance improves when replacing $s_{kpts}$ with a previously generated image of the subject, yielding better facial similarity while maintaining compositional control. 

In \textbf{face replacement}, users interact with a simple tool to manipulate (move/rotate/resize) their cropped face on a canvas matching the diffusion model's output dimensions. This approach eliminates the similarity issues caused by using another person's facial landmarks. However, the method performs poorly when none of the reference images show the subject facing the camera (deviations $>30\degree$). User satisfaction was higher with this approach compared to \textit{2-step generation}, which we attribute to increased interactivity and faster generation times.

\subsubsection{Augmentations}
Due to InstantID's architectural design, rotational and shape-altering augmentations proved ineffective. Background replacement and similar context-modifying augmentations degraded similarity because the resulting artifacts fall outside the distribution of images encountered during training by the model provided in \cite{Wang2024Jan}. The trained model demonstrates robustness to meaningful color adjustments, rendering color modifications unnecessary for well-lit scenes. For low-resolution images, traditional upscaling methods (Lanczos/Bicubic) performed adequately, while neural network-based upscaling introduced novel artifacts unseen during training, resulting in reduced quality.


\section{Survey}

We conducted the survey to evaluate the viability of AI-generated portraits for professional use and to compare the performance of DreamBooth and InstantID in generating realistic headshots. A pilot study was conducted prior to the main survey to ensure the clarity and relevance of the questions.
Numerical data on answers can be found in Suppl.~\ref{app:surveygraphs} and questionary can be found in Suppl.~\ref{app:surveyqs}.

A total of 97 participants, comprising current white-collar workers as well as students pursuing careers in white-collar fields, took part in the online survey.
The participant pool covered ages 18 to 74, with $12.4\%$ of respondents in the younger 18--24 bracket, but only $13.4\%$ in the 25--44 age range. Overall, the sample skewed older, with $74.2\%$ aged 45 or above. Gender was nearly evenly split between male and female, with no representation of diverse identities and several non-disclosures.
Among the respondents, 81 self-identified as White, 3 as Asian, 1 as Hispanic, and 12 chose not to disclose their ethnicity.
As the sample was predominantly White and older, findings may not fully generalize to younger or more diverse populations.


\textbf{Overall Performance of Generated Portraits}\quad    
Portraits generated by DreamBooth and InstantID performed similarly across multiple aspects, including overall quality, facial detail clarity, identity preservation, perceived level of editing, and background quality. Using high-quality subject datasets led to slightly better results in most categories, except for ``Editing,'' where participants indicated familiarity and acceptance of traditional Photoshop-enhanced portraits. 

\textbf{Method Preferences}\quad  
A slightly higher percentage of participants ($\approx$4\%) preferred the standardized portraits from InstantID over the more flexible outputs of DreamBooth. InstantID was often perceived as more professional, likely due to its consistent ``Photoshopped look,'' which resonated with a broader audience. Open-ended responses highlighted diverse preferences, with participants emphasizing factors such as lighting, pose, angle, expression, detail, color, and background.

\textbf{Facial Similarity}\quad  
DreamBooth demonstrated superior facial similarity between real images individuals and their generated portraits compared to InstantID. More participants identified InstantID images as depicting a different person than the reference. DreamBooth consistently maintained a higher level of facial similarity across both high- and low-quality subject datasets.

\textbf{Noticing AI Generations}\quad    
Most white-collar workers struggled to identify AI-generated headshots when not explicitly prompted, often focusing on well-known but absent flaws commonly associated with AI generation. Among a subset of participants $(n = 77)$ who regularly notice AI-generated images in daily life, the generated portraits blended well with conventional studio photographs. However, participants who actively use AI for image creation $(n = 29)$ demonstrated better identification skills. This group was more likely to recognize DreamBooth images as AI-generated, possibly due to DreamBooth's popularity, while InstantID generations, being more niche, had a near $50/50$ chance of being identified as AI.

\section{Discussion}

Our experiments offer insights into augmentation strategies for improving facial resemblance in personalized text-to-image generation using DreamBooth and InstantID. While classical augmentations are common in deep learning, applying them to few-shot personalization can yield undesirable results. Geometric transformations like flipping and rotation disrupted learning due to face asymmetry and artifacts. Color jittering caused erratic generations by associating color shifts with DreamBooth’s rare token. Background augmentations with U$^2$-Net introduced segmentation imperfections, especially around hair, which SDXL learned. Replacing backgrounds with patterns or studio backdrops also degraded image quality. Auto color adjustment with Adobe Lightroom improved color grading.

Generative augmentation via InstantID proved more effective for enhancing facial similarity in DreamBooth training. By generating diverse synthetic images with varied prompts and facial landmarks, we enriched the dataset with realistic examples, aligning it with the diffusion model’s space. However, maintaining a balance between real and InstantID-generated images is crucial to avoid overfitting and loss of recontextualization.

FaceDistance provided a quantitative measure of facial similarity but became less useful for hyperparameter tuning once a certain fidelity level was reached. A user survey among white-collar workers showed that both DreamBooth and InstantID performed similarly in quality, clarity, identity preservation, editing, and background. A slight preference emerged for the ``Photoshopped look'' of InstantID portraits. While DreamBooth achieved better facial similarity, many participants struggled to distinguish AI-generated images from real ones, particularly those unfamiliar with AI tools. Users actively engaged in AI image creation were more likely to identify DreamBooth images as synthetic, possibly due to its higher popularity.

InstantID’s effectiveness depends on reference image quality and diversity. Using multiple references (around four) improved similarity by enriching information for the Face Encoder. Facial landmarks strongly influenced pose and composition, sometimes overriding text prompts. We explored 2-step generation and interactive face replacement to enhance control, with the latter showing higher user satisfaction. Rotational and shape-altering augmentations were ineffective, and background modifications reduced similarity. Traditional upscaling worked well for low-resolution images, whereas neural network-based upscaling introduced artifacts.

\section{Limitations}

A key limitation is that InstantID-based augmentation reduces realism in generated images. While DreamBooth remains more flexible for personalized generation, InstantID-enhanced datasets still outperform unaugmented ones. Given the baseline model’s photorealism constraints, using generative augmentation to refine its training data is a practical approach.

We acknowledge that the survey sample is predominantly White and includes mostly older age brackets, which may introduce bias and limit the generalizability of the findings to more ethnically diverse and younger populations.

FaceDistance metric exhibits limited sensitivity, primarily supporting coarse discrimination between low- and high-quality generations rather than capturing nuanced, fine-grained differences. FaceDistance becomes less informative for hyperparameter tuning once generation quality surpasses a baseline level of fidelity. Furthermore, it does not effectively evaluate recontextualization or provide a holistic measure of personalization performance, limiting its utility in tasks where those dimensions are critical.

\section{Future Work}

Future work should improve FaceDistance’s sensitivity for subtle distinctions and expand its scope beyond facial similarity to include recontextualization and overall image quality. Integrating human preferences will better align evaluations with subjective aesthetics.

Given our goal of enhancing facial fidelity, we plan to adopt LPIPS \cite{Zhang2018Jan} in future evaluations, as it better captures perceptual differences relevant to visual realism. In contrast, we find DINO \cite{Caron2021Apr} is less suitable, emphasizing semantic similarity over identity or fine-grained fidelity.

Future work should consider ethical risks such as misuse, consent, identity theft, and the broader societal impact of personalized generation technologies.

\section{Conclusion}

This study examined augmentation strategies for improving facial resemblance in personalized image generation using DreamBooth and InstantID. Classical augmentations can introduce artifacts that degrade facial fidelity, requiring careful application.

We found generative augmentation with InstantID to be highly effective for improving DreamBooth training. Creating diverse, realistic synthetic images while maintaining a balanced ratio with real data prevents overfitting.

User surveys confirmed that both DreamBooth and InstantID produce high-quality, professional-looking headshots, often indistinguishable from real photos. While DreamBooth excels in facial similarity, InstantID’s consistent output appears more polished.

For practical implementation, we establish that using approximately four reference images optimizes InstantID performance, and techniques like 2-step generation and interactive face replacement enhance pose control and user satisfaction. These findings support the viability of generative augmentation as a practical approach for reducing dependency on real data in personalization tasks.

Overall, our findings provide insights into augmentation strategies for personalized image generation, guiding their application in tasks requiring high facial fidelity. Future work should explore advanced generative augmentation techniques and better user control over InstantID outputs.

{
    \small
    \bibliographystyle{ieeenat_fullname}
    \bibliography{main}

\begin{thebibliography}{28}
\providecommand{\natexlab}[1]{#1}
\providecommand{\url}[1]{\texttt{#1}}
\expandafter\ifx\csname urlstyle\endcsname\relax
  \providecommand{\doi}[1]{doi: #1}\else
  \providecommand{\doi}{doi: \begingroup \urlstyle{rm}\Url}\fi

\bibitem[Bib(2025{\natexlab{a}})]{BibEntry2025Ma54}
{ComfyUI{$\_$}InstantID}, 2025{\natexlab{a}}.
\newblock [Online; accessed 30. Mar. 2025].

\bibitem[Bib(2025{\natexlab{b}})]{BibEntry2025Mar2}
{SG161222/RealVisXL{$\_$}V4.0 {$\cdot$} Hugging Face}, 2025{\natexlab{b}}.
\newblock [Online; accessed 30. Mar. 2025].

\bibitem[Bib(2025{\natexlab{c}})]{BibEntry2025Mar234}
{TheBloke/dolphin-2.2.1-mistral-7B-GGUF {$\cdot$} Hugging Face}, 2025{\natexlab{c}}.
\newblock [Online; accessed 30. Mar. 2025].

\bibitem[Bib(2025{\natexlab{d}})]{BibEntry2025Mar3}
{sd-scripts}, 2025{\natexlab{d}}.
\newblock [Online; accessed 30. Mar. 2025].

\bibitem[wik(2025)]{wikiiiiii}
{Category:Patterns - Wikimedia Commons}, 2025.
\newblock [Online; accessed 31. Mar. 2025].

\bibitem[Agarwal et~al.(2022)Agarwal, Sen, Mukhopadhyay, Namboodiri, and Jawahar]{Agarwal2022Aug}
Aditya Agarwal, Bipasha Sen, Rudrabha Mukhopadhyay, Vinay Namboodiri, and C.~V. Jawahar.
\newblock {FaceOff: A Video-to-Video Face Swapping System}.
\newblock \emph{arXiv}, 2022.

\bibitem[Caron et~al.(2021)Caron, Touvron, Misra, J{\ifmmode\acute{e}\else\'{e}\fi}gou, Mairal, Bojanowski, and Joulin]{Caron2021Apr}
Mathilde Caron, Hugo Touvron, Ishan Misra, Herv{\ifmmode\acute{e}\else\'{e}\fi} J{\ifmmode\acute{e}\else\'{e}\fi}gou, Julien Mairal, Piotr Bojanowski, and Armand Joulin.
\newblock {Emerging Properties in Self-Supervised Vision Transformers}.
\newblock \emph{arXiv}, 2021.

\bibitem[Chang and Lekena(2024)]{GlamTry2024}
Ting-Yu Chang and Seretsi~Khabane Lekena.
\newblock {GlamTry: Advancing Virtual Try-On for High-End Accessories}.
\newblock \emph{arXiv}, 2024.

\bibitem[Islam and Akhtar(2025)]{Islam2025Mar}
Khawar Islam and Naveed Akhtar.
\newblock {Context-guided Responsible Data Augmentation with Diffusion Models}.
\newblock \emph{arXiv}, 2025.

\bibitem[Islam et~al.(2024)Islam, Zaheer, Mahmood, and Nandakumar]{Islam2024Apr}
Khawar Islam, Muhammad~Zaigham Zaheer, Arif Mahmood, and Karthik Nandakumar.
\newblock {DiffuseMix: Label-Preserving Data Augmentation with Diffusion Models}.
\newblock \emph{arXiv}, 2024.

\bibitem[Karras et~al.(2018)Karras, Laine, and Aila]{Karras2018Dec}
Tero Karras, Samuli Laine, and Timo Aila.
\newblock {A Style-Based Generator Architecture for Generative Adversarial Networks}.
\newblock \emph{arXiv}, 2018.

\bibitem[Kazemi and Sullivan(2014)]{Kazemi2014Jun}
Vahid Kazemi and Josephine Sullivan.
\newblock {One millisecond face alignment with an ensemble of regression trees}.
\newblock \emph{ResearchGate}, pages 1867--1874, 2014.

\bibitem[Li et~al.(2020)Li, Zhang, Maybank, and Tao]{Li2020Oct}
Jizhizi Li, Jing Zhang, Stephen~J. Maybank, and Dacheng Tao.
\newblock {Bridging Composite and Real: Towards End-to-end Deep Image Matting}.
\newblock \emph{arXiv}, 2020.

\bibitem[Li et~al.(2019)Li, Bao, Yang, Chen, and Wen]{Li2019Dec}
Lingzhi Li, Jianmin Bao, Hao Yang, Dong Chen, and Fang Wen.
\newblock {FaceShifter: Towards High Fidelity And Occlusion Aware Face Swapping}.
\newblock \emph{arXiv}, 2019.

\bibitem[Liew et~al.(2022)Liew, Yan, Zhou, and Feng]{Liew2022Oct}
Jun~Hao Liew, Hanshu Yan, Daquan Zhou, and Jiashi Feng.
\newblock {MagicMix: Semantic Mixing with Diffusion Models}.
\newblock \emph{arXiv}, 2022.

\bibitem[Nichol and Dhariwal(2021)]{Nichol2021Feb}
Alex Nichol and Prafulla Dhariwal.
\newblock {Improved Denoising Diffusion Probabilistic Models}.
\newblock \emph{arXiv}, 2021.

\bibitem[Podell et~al.(2023)Podell, English, Lacey, Blattmann, Dockhorn, M{\ifmmode\ddot{u}\else\"{u}\fi}ller, Penna, and Rombach]{Podell2023Jul}
Dustin Podell, Zion English, Kyle Lacey, Andreas Blattmann, Tim Dockhorn, Jonas M{\ifmmode\ddot{u}\else\"{u}\fi}ller, Joe Penna, and Robin Rombach.
\newblock {SDXL: Improving Latent Diffusion Models for High-Resolution Image Synthesis}.
\newblock \emph{arXiv}, 2023.

\bibitem[Qin et~al.(2020)Qin, Zhang, Huang, Dehghan, Zaiane, and Jagersand]{Qin2020May}
Xuebin Qin, Zichen Zhang, Chenyang Huang, Masood Dehghan, Osmar~R. Zaiane, and Martin Jagersand.
\newblock {U$^2$-Net: Going Deeper with Nested U-Structure for Salient Object Detection}.
\newblock \emph{arXiv}, 2020.

\bibitem[Rombach et~al.(2022)Rombach, Blattmann, Lorenz, Esser, and Ommer]{Rombach2022}
Robin Rombach, Andreas Blattmann, Dominik Lorenz, Patrick Esser, and Bj{\ifmmode\ddot{o}\else\"{o}\fi}rn Ommer.
\newblock {High-Resolution Image Synthesis With Latent Diffusion Models}, 2022.
\newblock [Online; accessed 31. Mar. 2025].

\bibitem[Ruiz et~al.(2022)Ruiz, Li, Jampani, Pritch, Rubinstein, and Aberman]{Ruiz2022Aug}
Nataniel Ruiz, Yuanzhen Li, Varun Jampani, Yael Pritch, Michael Rubinstein, and Kfir Aberman.
\newblock {DreamBooth: Fine Tuning Text-to-Image Diffusion Models for Subject-Driven Generation}.
\newblock \emph{arXiv}, 2022.

\bibitem[Schroff et~al.(2015)Schroff, Kalenichenko, and Philbin]{Schroff2015Mar}
Florian Schroff, Dmitry Kalenichenko, and James Philbin.
\newblock {FaceNet: A Unified Embedding for Face Recognition and Clustering}.
\newblock \emph{arXiv}, 2015.

\bibitem[Serengil and {\ifmmode\ddot{O}\else\"{O}\fi}zp{\ifmmode\imath\else\i\fi}nar(2024)]{Serengil2024Mar}
Sefik Serengil and Alper {\ifmmode\ddot{O}\else\"{O}\fi}zp{\ifmmode\imath\else\i\fi}nar.
\newblock {A Benchmark of Facial Recognition Pipelines and Co-Usability Performances of Modules}.
\newblock \emph{Bili{\ifmmode\mbox{\c{s}}\else\c{s}\fi}im Teknolojileri Dergisi}, 17\penalty0 (2):\penalty0 95--107, 2024.

\bibitem[Trabucco et~al.(2023)Trabucco, Doherty, Gurinas, and Salakhutdinov]{Trabucco2023Feb}
Brandon Trabucco, Kyle Doherty, Max Gurinas, and Ruslan Salakhutdinov.
\newblock {Effective Data Augmentation With Diffusion Models}.
\newblock \emph{arXiv}, 2023.

\bibitem[Wang et~al.(2024)Wang, Bai, Qin, Chen, Li, Tang, and Hu]{Wang2024Jan}
Qixun Wang, Xu Bai, Zekui Qin, Anthony Chen, Huaxia Li, Xu Tang, and Yao Hu.
\newblock {InstantID: Zero-shot Identity-Preserving Generation in Seconds}.
\newblock \emph{arXiv}, 2024.

\bibitem[Wang et~al.(2018)Wang, Yu, Wu, Gu, Liu, and Dong]{Wang2018Sep}
Xintao Wang, Ke Yu, Shixiang Wu, Jinjin Gu, Yihao Liu, and Chao Dong.
\newblock {ESRGAN: Enhanced Super-Resolution Generative Adversarial Networks}.
\newblock \emph{arXiv}, 2018.

\bibitem[{Wikimedia Commons contributors}(2025)]{wikimedia_patterns}
{Wikimedia Commons contributors}.
\newblock Category: Patterns.
\newblock \url{https://commons.wikimedia.org/wiki/Category:Patterns}, 2025.
\newblock Accessed: 2025-06-19.

\bibitem[Zhang et~al.(2016)Zhang, Zhang, Li, and Qiao]{Zhang2016Apr}
Kaipeng Zhang, Zhanpeng Zhang, Zhifeng Li, and Yu Qiao.
\newblock {Joint Face Detection and Alignment using Multi-task Cascaded Convolutional Networks}.
\newblock \emph{arXiv}, 2016.

\bibitem[Zhang et~al.(2018)Zhang, Isola, Efros, Shechtman, and Wang]{Zhang2018Jan}
Richard Zhang, Phillip Isola, Alexei~A. Efros, Eli Shechtman, and Oliver Wang.
\newblock {The Unreasonable Effectiveness of Deep Features as a Perceptual Metric}.
\newblock \emph{arXiv}, 2018.

\end{thebibliography}
}

\clearpage
\setcounter{page}{1}

\maketitlesupplementary

\section{FaceDistance in Practice}
\label{app:facedistance-downstream-apps}

In downstream applications such as AI-generated portraits, it is common practice to generate large batches of images ($200+$) to ensure sufficient variety and quality. However, this brute-force approach places a significant burden on users, who must manually review and select from hundreds of samples, leading to inefficiency and cognitive overload. 

By leveraging FaceDistance to filter and rank images based on their similarity to the reference face, we can greatly reduce this friction. This enables early stopping once enough high-quality matches are found or allows prioritization of the top-ranked images, streamlining the selection process and enhancing overall user satisfaction.

\begin{figure}[bht]
    \centering
    \includegraphics[width=0.98\linewidth]{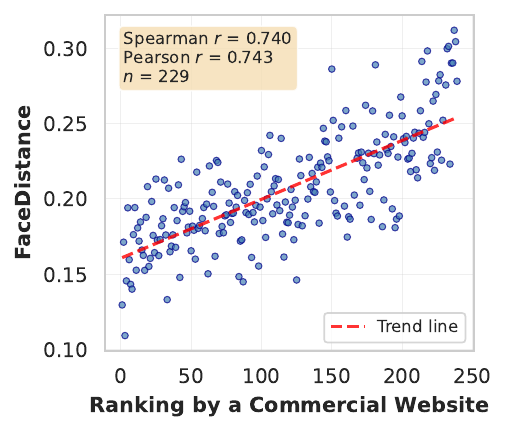}
    \caption{Correlation between FaceDistance and ranking from a commercial website for 229 AI-generated portrait samples. Due to proprietary considerations, we do not disclose the identity of the commercial website.}
    \label{fig:face_distance_vs_original_number}
\end{figure}

FaceDistance rankings exhibit a strong correlation with those from commercial platforms (Figure~\ref{fig:face_distance_vs_original_number}), demonstrating that our metric effectively captures perceived facial similarity. This suggests FaceDistance captures a fundamental aspect of facial similarity that underlies the ranking criteria used by these platforms, highlighting its novelty and potential as a core component in automated facial assessment.

\begin{figure*}
    
    \centering
    \setlength{\tabcolsep}{3pt} 
    \renewcommand{\arraystretch}{1.5} 
    \begin{tabular}{c|ccccc}
        \textbf{Real} & \textbf{Min (Best)} & \textbf{Q1 (0.182)} & \textbf{Q2 (0.200)} & \textbf{Q3 (0.229)} & \textbf{Max (Worst)} \\

        \includegraphics[width=0.15\linewidth]{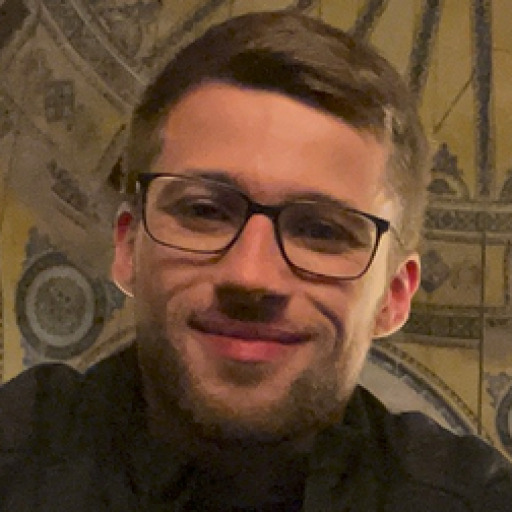} & 
        \includegraphics[width=0.15\linewidth]{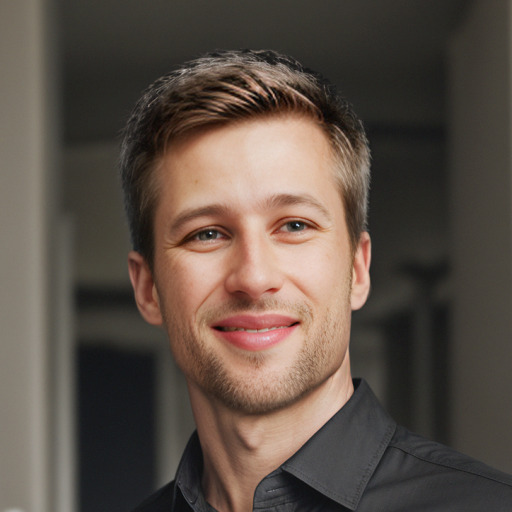} & 
        \includegraphics[width=0.15\linewidth]{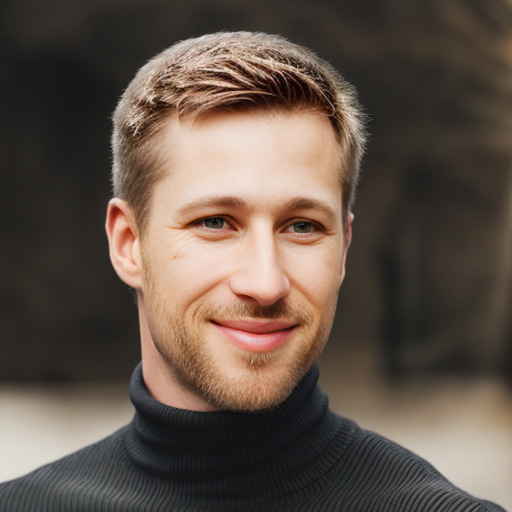} & 
        \includegraphics[width=0.15\linewidth]{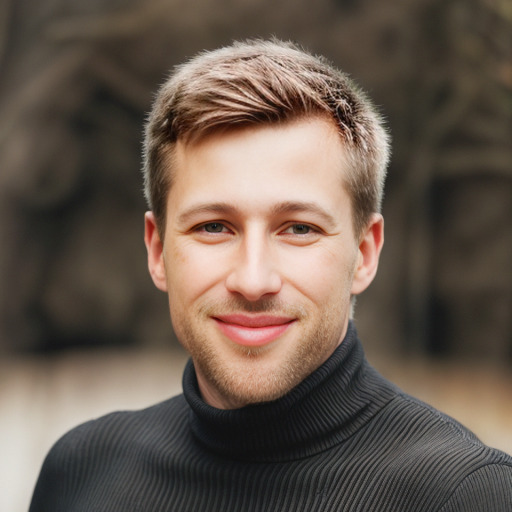} & 
        \includegraphics[width=0.15\linewidth]{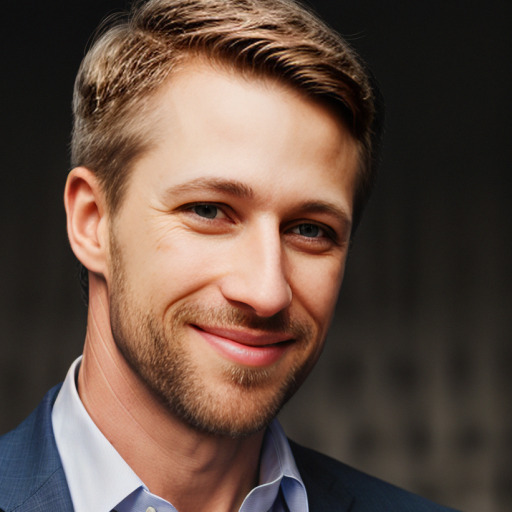} & 
        \includegraphics[width=0.15\linewidth]{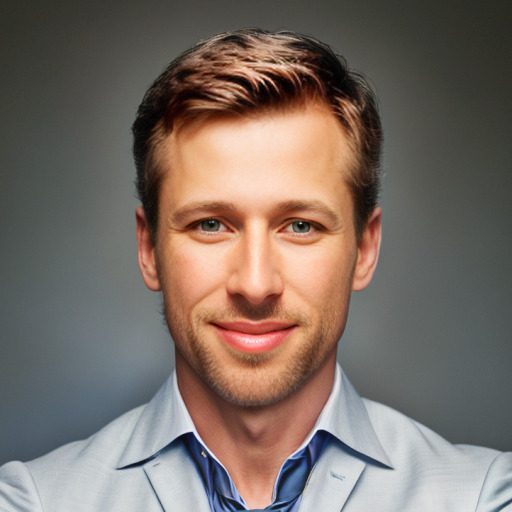} \\

        \includegraphics[width=0.15\linewidth]{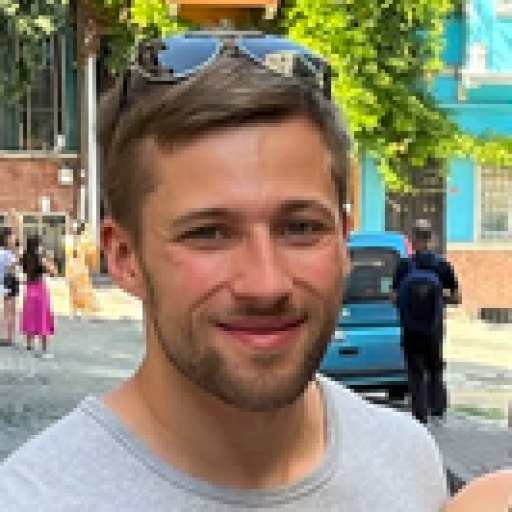} & 
        \includegraphics[width=0.15\linewidth]{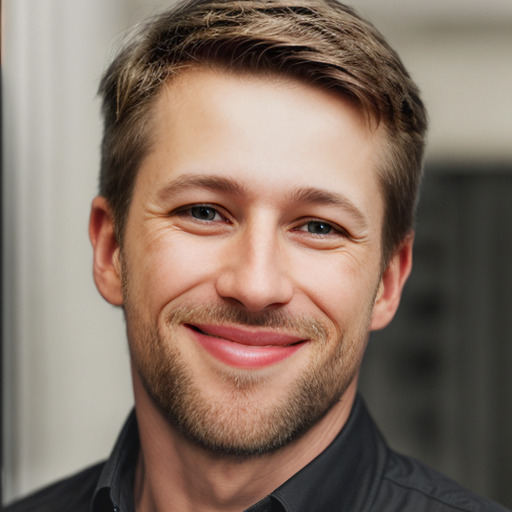} & 
        \includegraphics[width=0.15\linewidth]{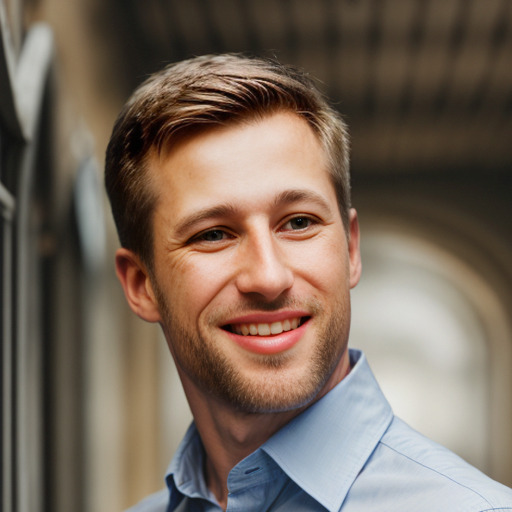} & 
        \includegraphics[width=0.15\linewidth]{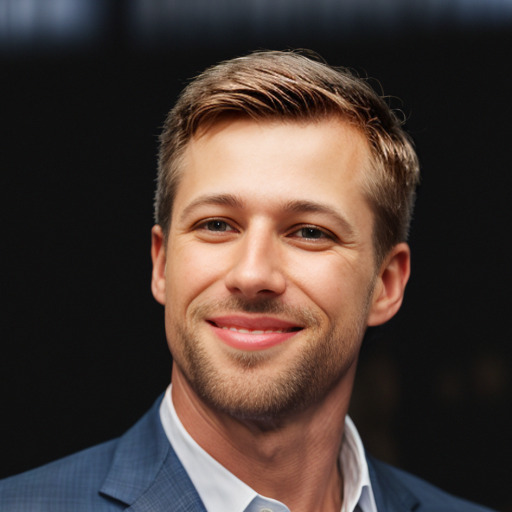} & 
        \includegraphics[width=0.15\linewidth]{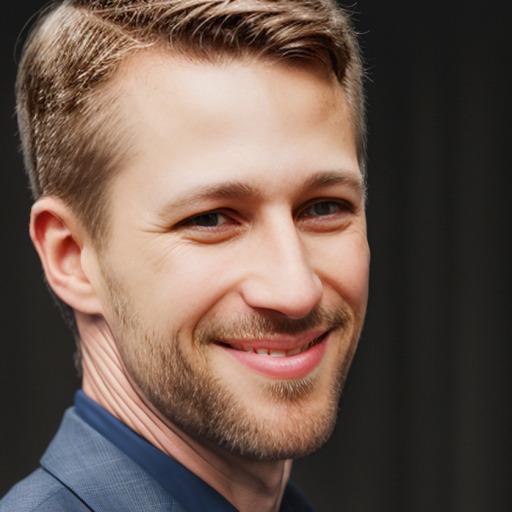} & 
        \includegraphics[width=0.15\linewidth]{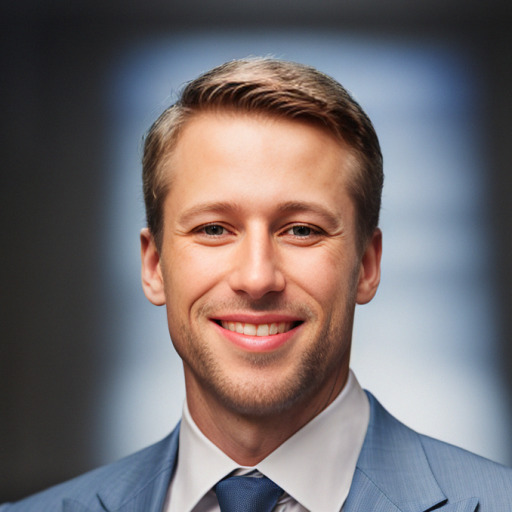} \\

        \includegraphics[width=0.15\linewidth]{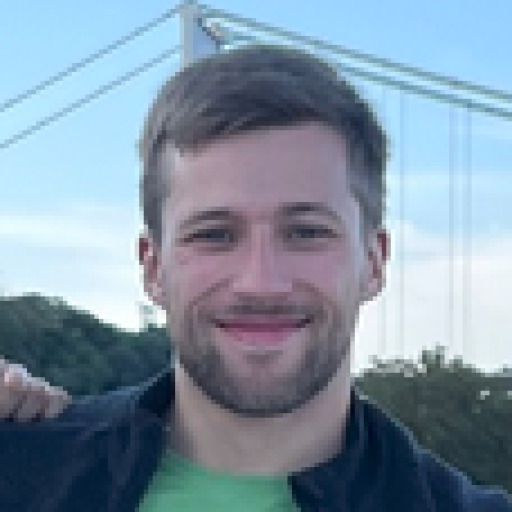} & 
        \includegraphics[width=0.15\linewidth]{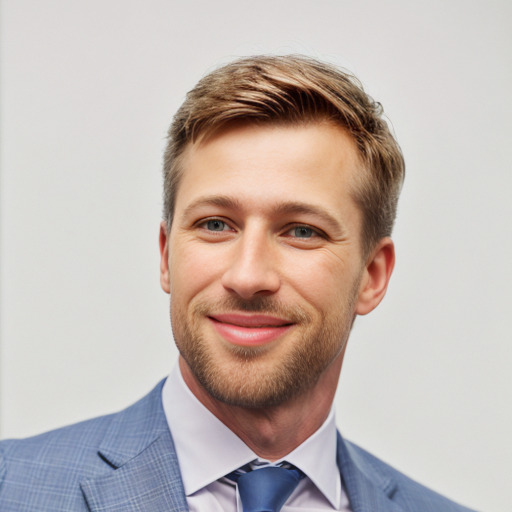} & 
        \includegraphics[width=0.15\linewidth]{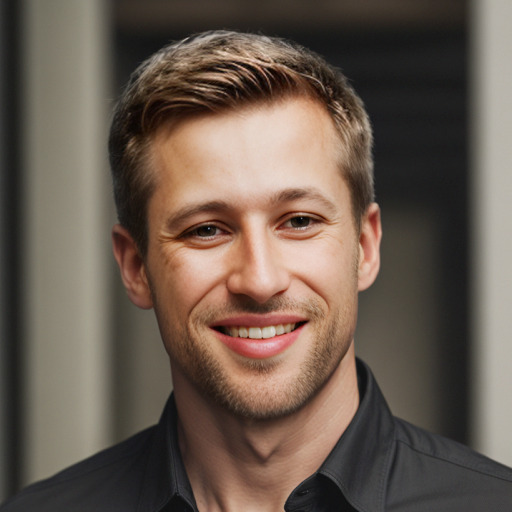} & 
        \includegraphics[width=0.15\linewidth]{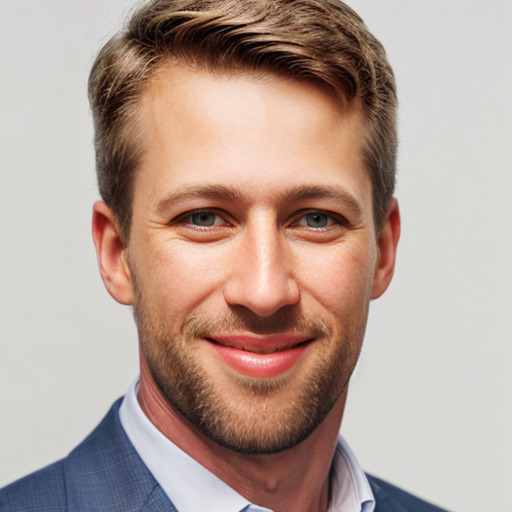} & 
        \includegraphics[width=0.15\linewidth]{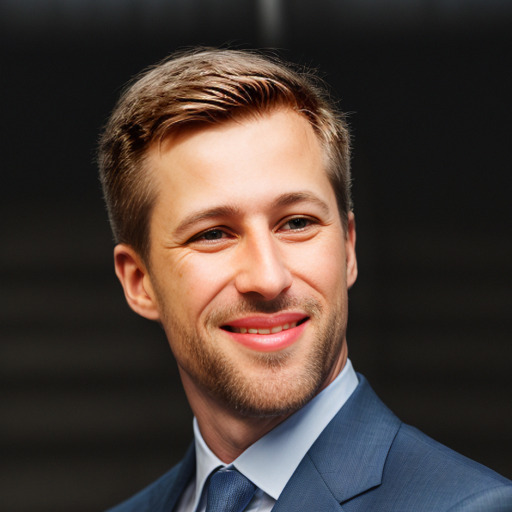} & 
        \includegraphics[width=0.15\linewidth]{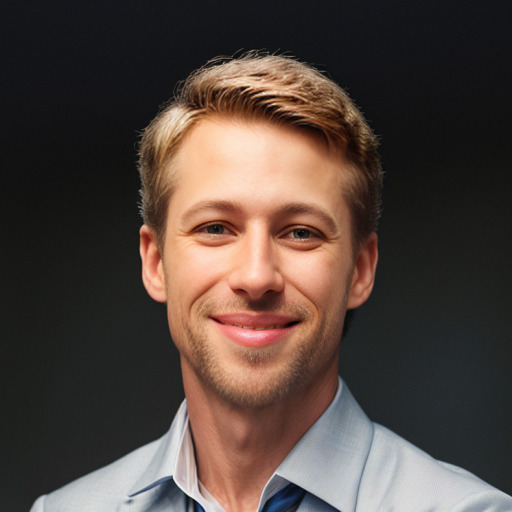} \\
    \end{tabular}
    \caption{
       Visual Comparison of Real and Generated Faces Across Similarity Bins.
        Columns represent increasing face distance (left to right), while rows display varied samples. Faces are sorted by similarity using FaceDistance, ranging from 0.109 to 0.312.
    }
    \label{fig:facedistance-downstream-apps}
\end{figure*}

Figure~\ref{fig:facedistance-downstream-apps} further shows how samples grouped by FaceDistance correspond to visual likeness, demonstrating FaceDistance as a reliable and interpretable metric for selection that enhances efficiency and satisfaction.

\section{FaceDistance of InstantID generations}
\label{app:facedistance-img-count}

\begin{figure}[htb]
\centering
\includegraphics[width=0.98\linewidth]{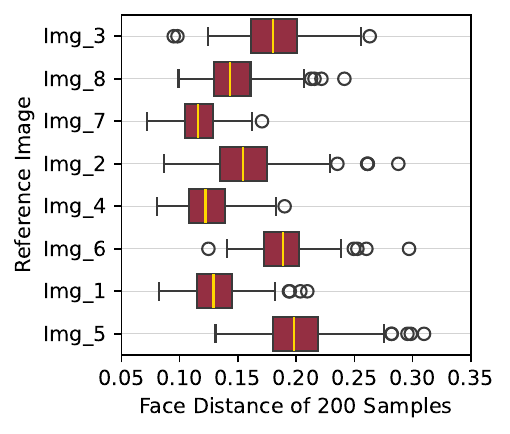}
\caption{Distribution of facial similarity metrics for 8 different input subjects using single reference images, with the y-axis sorted by FaceDistance (meaning Img\_3 has the lowest FaceDistance within, Img\_5 has the highest).}
\label{fig:k1}
\end{figure}

\begin{figure}[htb]
\centering
\includegraphics[width=0.98\linewidth]{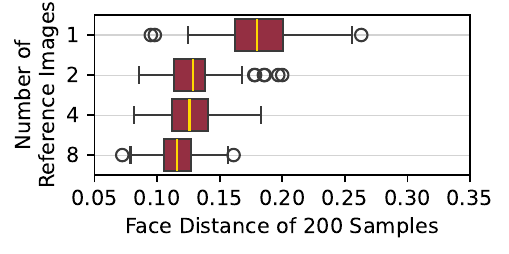}
\caption{Comparison of facial similarity performance when using k reference images, demonstrating the leftward shift in FaceDistance distribution as $k$ increases from 1 to 8 images.}
\label{fig:k2}
\end{figure}

Our experiments with InstantID show that using a single reference image leads to high variance in output quality and inconsistent facial similarity (Figure~\ref{fig:k1}). In contrast, multiple reference images significantly improve facial consistency (Figure~\ref{fig:k2}). Analysis of the \textit{Vacation-Anna} dataset reveals that while the improvement from 1 to 2 images is substantial, the differences between 2, 4, and 8 images are subtle and often imperceptible without direct side-by-side comparison. The FaceDistance metric supports this, with the distribution shifting leftward as more reference images are used, indicating greater overall similarity.

\begin{figure}[htb]
\centering
\includegraphics[width=0.98\linewidth]{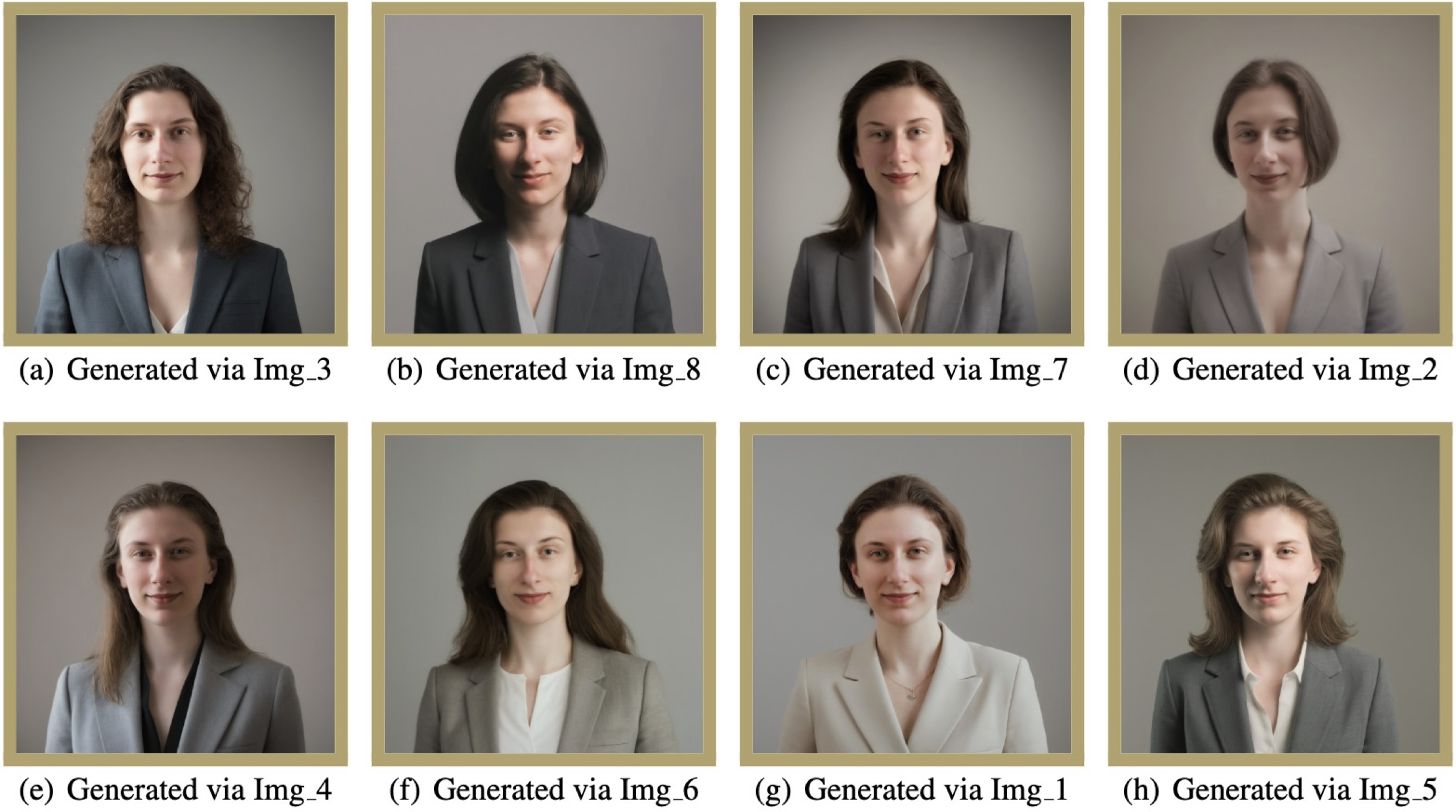}
\caption{Illustrative Images of Figure~\ref{fig:k1}. Generated using the 2-step method (Section~\ref{sec:landmarks-img}).}
\label{fig:sample-instantid-generations-s}
\end{figure}

\section{Motivating Alternatives to InstantID}

\begin{figure}[phtb]
    \centering
    \includegraphics[width=0.98\linewidth]{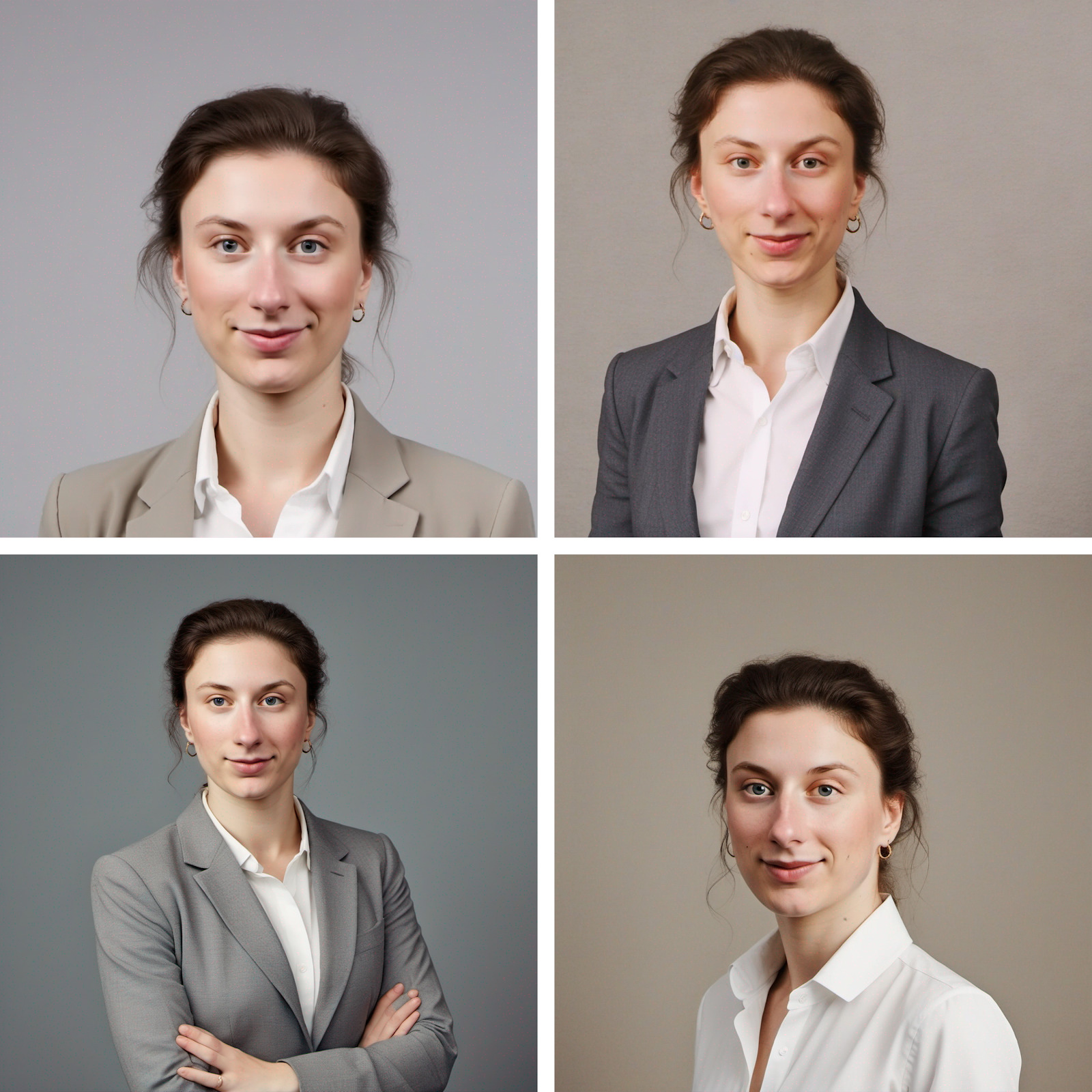}
    \caption{Dreambooth generated images with more variability \textit{(Vacation-Anna)}}
    \label{fig:db1}
\end{figure}

\begin{figure}[phtb]
    \centering
    \includegraphics[width=0.98\linewidth]{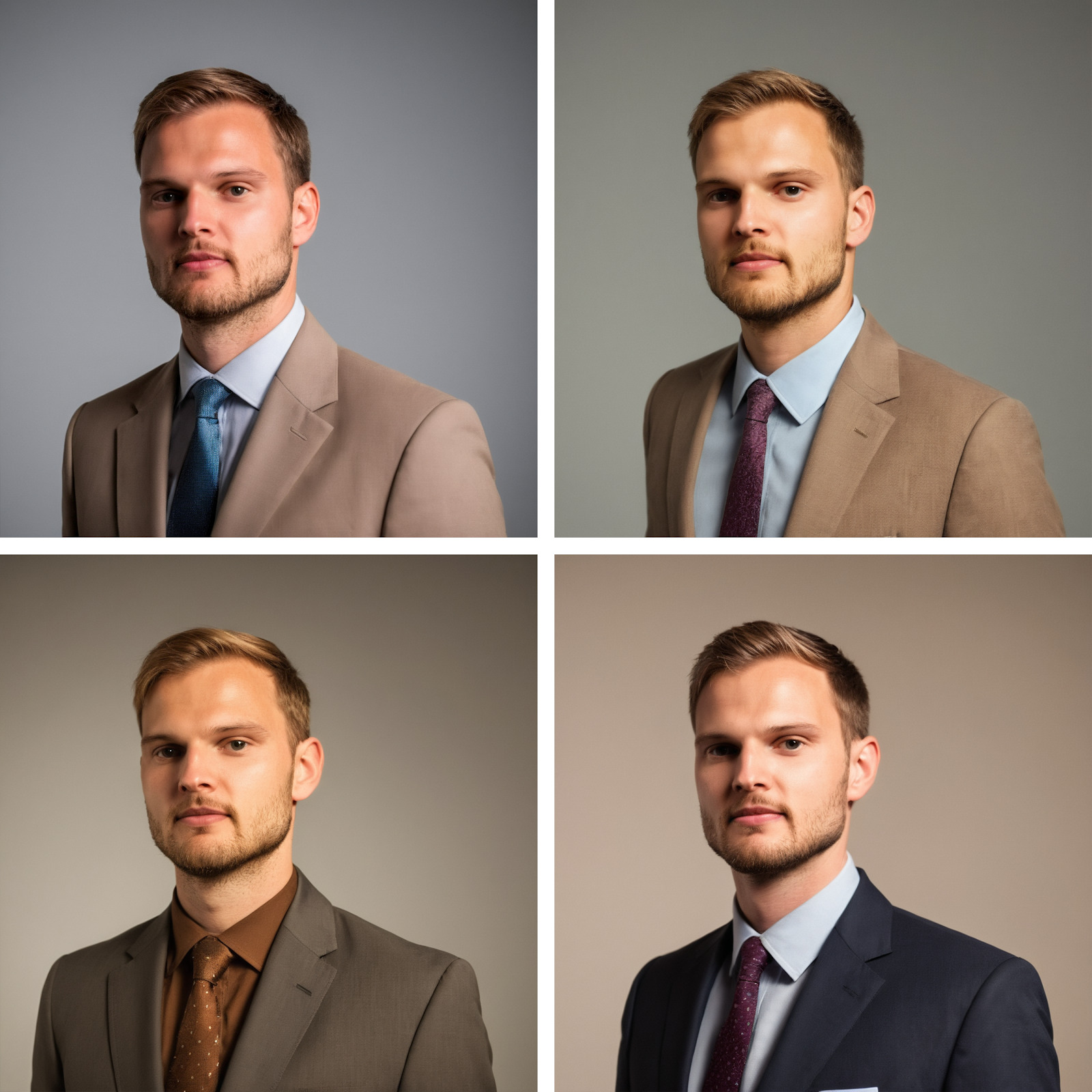}
    \caption{InstantID generated images with better facial similarity but with less variability \textit{(Party-Ryan)}}
    \label{fig:inid1}
\end{figure}

Although InstantID produces visually appealing images, its output variability for identical configurations is limited (see Figure~\ref{fig:inid1}). Our survey reveals that users are seeking greater diversity in images while maintaining a natural appearance, free from a photoshopped aesthetic. In contrast, utilizing Dreambooth for image generation circumvents this issue, making it the more favorable option (Figure~\ref{fig:db1}).

\section{Comparative Survey Analysis of Portrait Generation Methods}

\label{app:surveygraphs}

\begin{figure}[tb]
    \centering
    \includegraphics[width=0.98\linewidth]{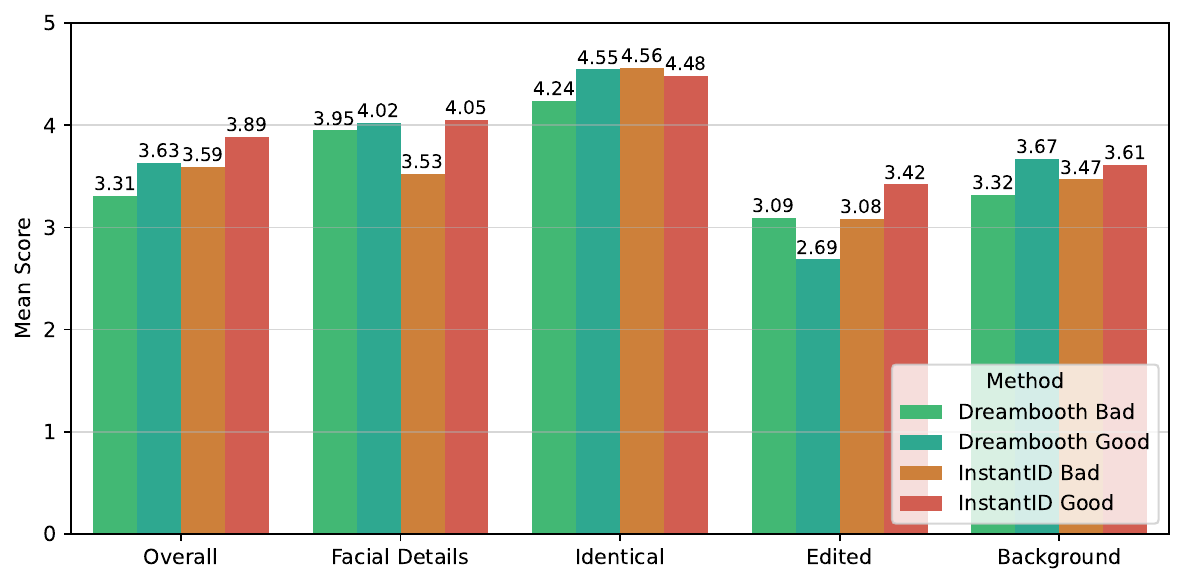}
    \caption{Mean Scores by Category for Each Method. Users were asked to rate samples from each method on 5 Categories, 5 being the highest value.}
    \label{fig:s1}
\end{figure}

\begin{figure}[tb]
    \centering
    \includegraphics[width=0.98\linewidth]{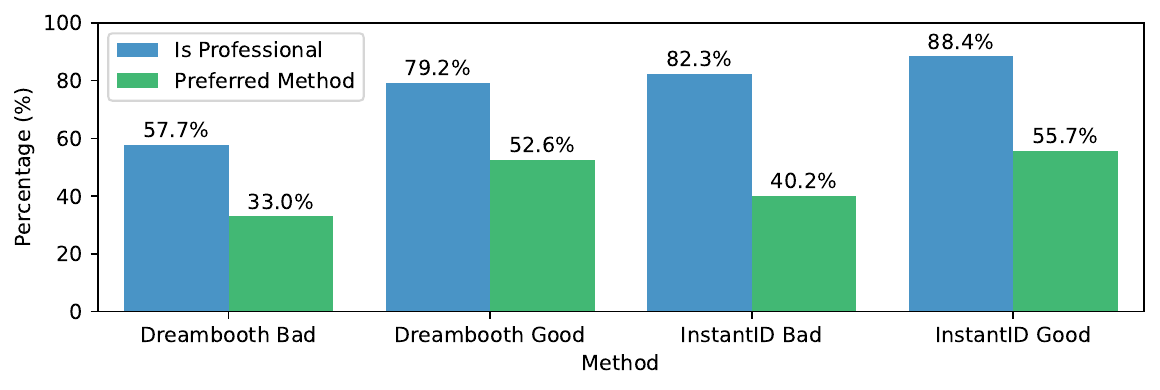}
    \caption{User Study Results: Participant Choices for 'Is Professional?' and 'Preferred Method' per AI Generation Technique.}
    \label{fig:s2}
\end{figure}

\begin{figure}[tb]
    \centering
    \includegraphics[width=0.98\linewidth]{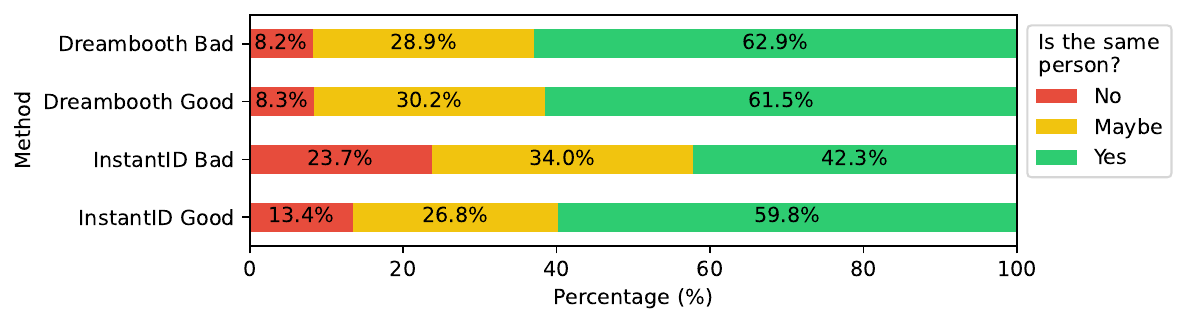}
    \caption{User Study Results: Participant Similarity Ratings ('Is the Same Person?') for Generated Images by Method}
    \label{fig:s3}
\end{figure}

Our analysis reveals that generated portraits perform similarly across individual aspects, likely due to the SD model employed (Figure~\ref{fig:s1}). Good subject datasets for Dreambooth yield more natural results, though user preferences trend toward Photoshop-edited appearances. Further investigation shows that ``good'' datasets correlate with higher employee agreement on preferred methods, with approximately 4\% more participants favoring standardized InstantID portraits for their professional, photoshopped aesthetic (Figure~\ref{fig:s2}). When assessing facial similarity between real and generated images, Dreambooth consistently outperformed InstantID regardless of dataset quality, with fewer participants identifying Dreambooth generations as different individuals (Figure~\ref{fig:s3}).

\begin{figure}[tb]
    \centering
    \includegraphics[width=0.98\linewidth]{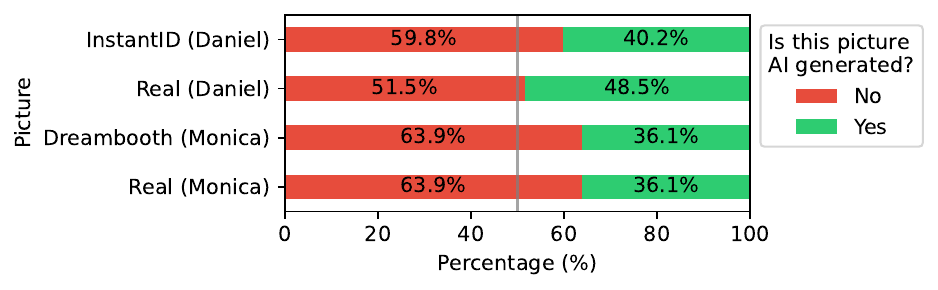}
    \caption{User Study Results: Participant Responses to 'Is This Picture AI Generated?' per Image $(n = 77)$}
    \label{fig:s4}
\end{figure}

\begin{figure}[tb]
    \centering
    \includegraphics[width=0.98\linewidth]{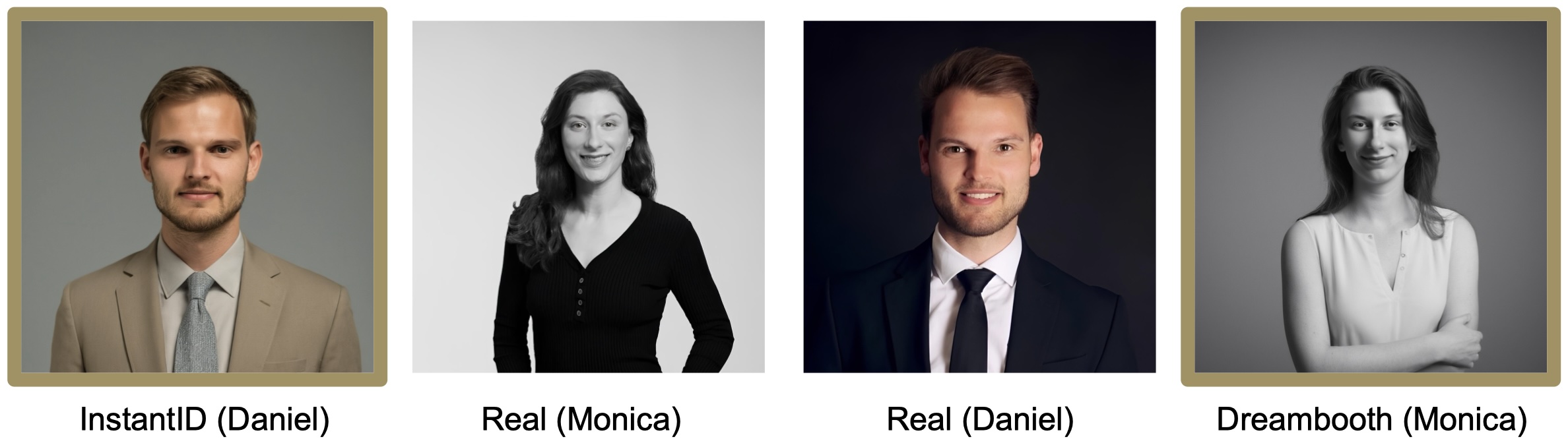}
    \caption{Sample images from Real/Fake Questions. (Results in Figure~\ref{fig:s4})}
    \label{fig:real-fake-detection-qs}
\end{figure}

White-collar workers demonstrated limited ability to distinguish AI-generated images, often searching for well-known AI flaws absent from our high-quality generations. Among the subset of participants ($n = 77$) who reported noticing AI-generated images in daily life, our generations integrated seamlessly with non-generated studio images, with only the ``Real (Daniel)'' image—featuring strong contrast and sharp features—dividing opinion nearly equally (Figure~\ref{fig:real-fake-detection-qs}, Figure~\ref{fig:s4}).


\section{Survey Questionnaire}
\label{app:surveyqs}

The following is the complete questionnaire used in our user study. Answer options are shown in \textit{italic}, while section titles and key instructions are marked in \textbf{bold}. Where applicable, questions were paired with image descriptions to provide visual context.

\begin{enumerate}
    \item[] \textbf{Title: Survey on Studio Portraits} \\
    \textit{Brief introductory text shown to participants.}
    
    \item \textbf{Rate the skill level of the photographer} \\
    \textit{((This section was repeated for each of four subjects. Each subject had four generated portraits shown in a grid (e.g., Figure~\ref{fig:db1}, Figure~\ref{fig:inid1}) along with a brief description of their professional background (e.g., ``marketer'', ``researcher'', ``nurse''). Use of AI is not disclosed to participants, yet.))}
    \begin{enumerate}
        \item How would you rate the overall quality?\\
        \makebox[\linewidth][r]{\textit{[Really Bad] 1--5 [Really Good]}}

        \item Are the facial details clear and well-defined?\\
        \makebox[\linewidth][r]{\textit{[No] 1--5 [Yes]}}

        \item How identical is the person in these pictures?\\
        \makebox[\linewidth][r]{\textit{[Completely Different] 1--5 [Exactly the Same]}}

        \item How much editing, if any, is present in this photo?\\
        \makebox[\linewidth][r]{\textit{[No Editing] 1--5 [Heavily Edited]}}

        \item Rate the quality of the background in the headshot.\\
        \makebox[\linewidth][r]{\textit{[Poor] 1--5 [Excellent]}}

        \item Would you expect to see these photos in a professional context (e.g., LinkedIn)?
        \hfill\textit{[Yes/No]}

        \item Is there anything you dislike about these pictures?\\
        \makebox[\linewidth][r]{\textit{[Text Answer]}}

        \item Is there anything you particularly like about these pictures?
        \hfill\textit{[Text Answer]}
    \end{enumerate}

    \item \textbf{Photographer Preference} \\
    \textit{If you liked at least one image, select the square near it.} \\
    \makebox[\linewidth][r]{\textit{[Four image grids from above]}}

    \item \textbf{Similarity of Real-Life Pictures and Portraits} \\
    \textit{Compare how similar the person in the everyday photo (left) is to the person in the portrait (right).}\\
    How would you rate the similarity of the person?\\
    \makebox[\linewidth][r]{\textit{[Not the Same] 1--3 [The Same]}}

    \item \textbf{Detection of AI Use} \\
    \textit{Some images may have been created using AI. Do not revise earlier responses.}\\
    Please select any photos you believe were AI-generated.\\
    \makebox[\linewidth][r]{\textit{[Four subject images]}}

    \item What influenced your choice?
    \hfill\textit{[Text Answer]}

    \item \textbf{Participant Background: Photography Experience}
    \begin{enumerate}
        \item How often do you take photos of yourself or others?\\
        \textit{[Daily or Weekly, Monthly, Every Few Months, Yearly or Never]}

        \item Do you use any software to edit your photos?\\
        \makebox[\linewidth][r]{\textit{[Yes/No]}}

        \item Do you frequently encounter AI-generated images in daily life?
        \hfill\textit{[Yes/No]}

        \item Have you ever used AI tools to create images?\\
        \makebox[\linewidth][r]{\textit{[Yes/No]}}
    \end{enumerate}
\end{enumerate}

We thank all participants for their time and contributions.

\section{Flickr-Suits-XL and Flickr-Portraits-XL Datasets} 
\label{sec:fsxl}

\begin{figure}[htb]
    \centering
    \includegraphics[width=0.98\linewidth]{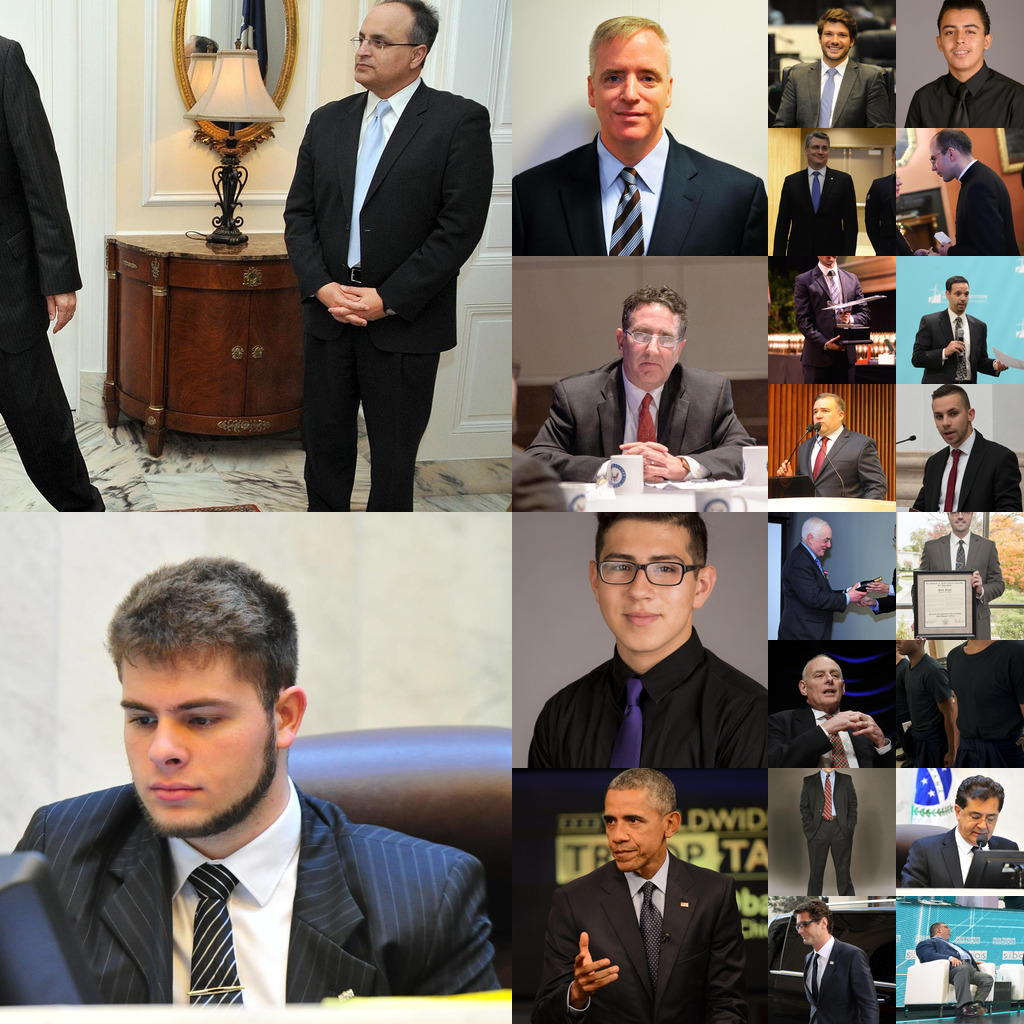}
    \caption{Samples from the Flickr-Suits-XL (FSXL) dataset. Images are center-cropped in this preview.}
    \label{fig:fsxl_examples}
\end{figure}

We have collected a new dataset of people in suits, Flickr-Suits-XL (FSXL), consisting of 1208 high-quality images (Figure~\ref{fig:fsxl_examples}). A noticeable gender imbalance is present in the FSXL dataset, with male subjects being the majority.

These images were obtained by crawling Flickr for 5687 images (thus inheriting all the biases of that website) under permissive licenses. We kept images with 1 visible face with MTCNN \cite{Zhang2016Apr}, aligned \cite{Kazemi2014Jun} and cropped the images (to the closest aspect ratio in SDXL training resolutions \cite{Podell2023Jul} i.e. $\approx 1$MP), maintaining the detected face centered horizontally and $\frac{1}{3}$ from top vertically. 

We have made the dataset publicly available at \\\href{https://github.com/KorayUlusan/fsxl-dataset}{https://github.com/KorayUlusan/fsxl-dataset}

\begin{figure}[htb]
    \centering
    \includegraphics[width=0.98\linewidth]{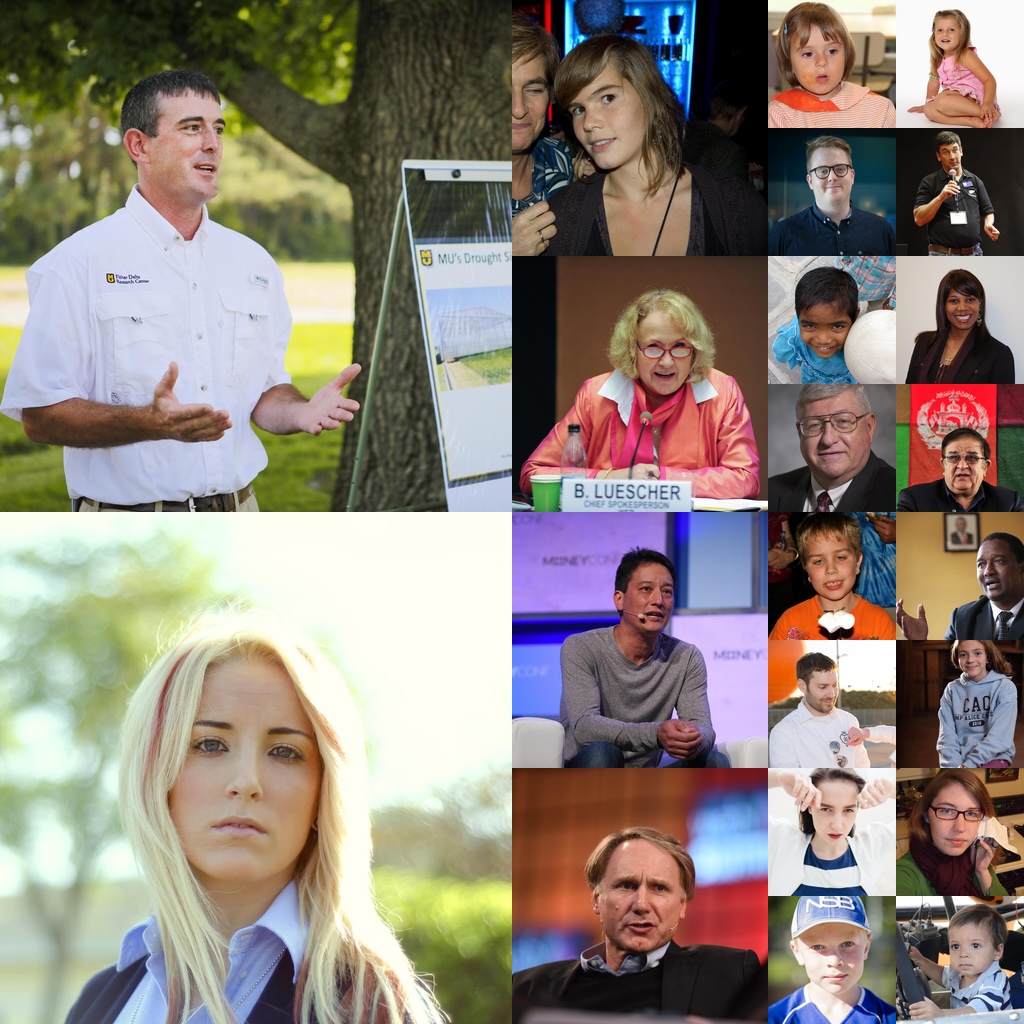}
    \caption{Samples from the Flickr-Portraits-XL (FPXL) dataset. Images are center-cropped in this preview (making it look like FFHQ).}
    \label{fig:fpxl_examples}
\end{figure}

Flickr-Portraits-XL (FPXL) follows the same procedure as FSXL but uses ``in-the-wild'' raw images of FFHQ dataset \cite{Karras2018Dec}, inheriting all the biases. (Figure~\ref{fig:fpxl_examples})
We thank the authors of the FFHQ dataset.

We have made the dataset publicly available at \\\href{https://github.com/KorayUlusan/fpxl-dataset}{https://github.com/KorayUlusan/fpxl-dataset}

One of the motivations in writing this paper was to create LinkedIn style portraits. Collecting such images was a natural step forward.

\section{Investigating Alternative Regularization for DreamBooth: A Case Study on Reconfirming Class Prior Preservation Principles}

DreamBooth \cite{Ruiz2022Aug} effectively personalizes large text-to-image diffusion models, enabling the synthesis of novel renditions of specific subjects from just a few input images. A critical element for its success is the autogenous class-specific prior preservation loss (PPL). This loss is specifically designed to counteract two common issues during fine-tuning: ``language drift,'' where the model forgets how to generate subjects of the target class, and the risk of reduced output diversity. DreamBooth mitigates these by supervising the model with its \textit{own generated samples}, thereby leveraging and retaining the semantic prior embedded within the model itself. This approach is crucial for preventing overfitting and ensuring the personalized subject can be generated in diverse contexts, poses, and views while maintaining its identity and the broader class knowledge.

In our study, complementing our primary research on augmentations, we investigated an alternative regularization strategy: utilizing external image collections instead of the model's self-generated samples for prior preservation. We used Flickr-Portraits-XL (FPXL) and and Flickr-Suits-XL (FSXL) datasets (Appendix~\ref{sec:fsxl}) to asses whether they can serve as regularization images to preserve the class prior during DreamBooth fine-tuning, akin to the objective of DreamBooth's PPL.


However, our experiments revealed a significant and unexpected outcome: using FPXL and FSXL as regularization datasets fundamentally destroyed the class word prior. After only a few training steps, the DreamBooth model could no longer properly generate images of ``a [class word]'' (e.g., ``a man'' or ``a woman''). This failure occurred because the distributions of these external datasets diverged significantly from the Stable Diffusion model’s inherent prior for the class word. Additionally, differences in camera field of view and cropping in FPXL and FSXL images—compared to the self-generated images typically used for regularization—exacerbated this divergence, especially with a fast-converging learning rate.

\begin{figure}[h!]
    \centering
    \includegraphics[width=0.98\linewidth]{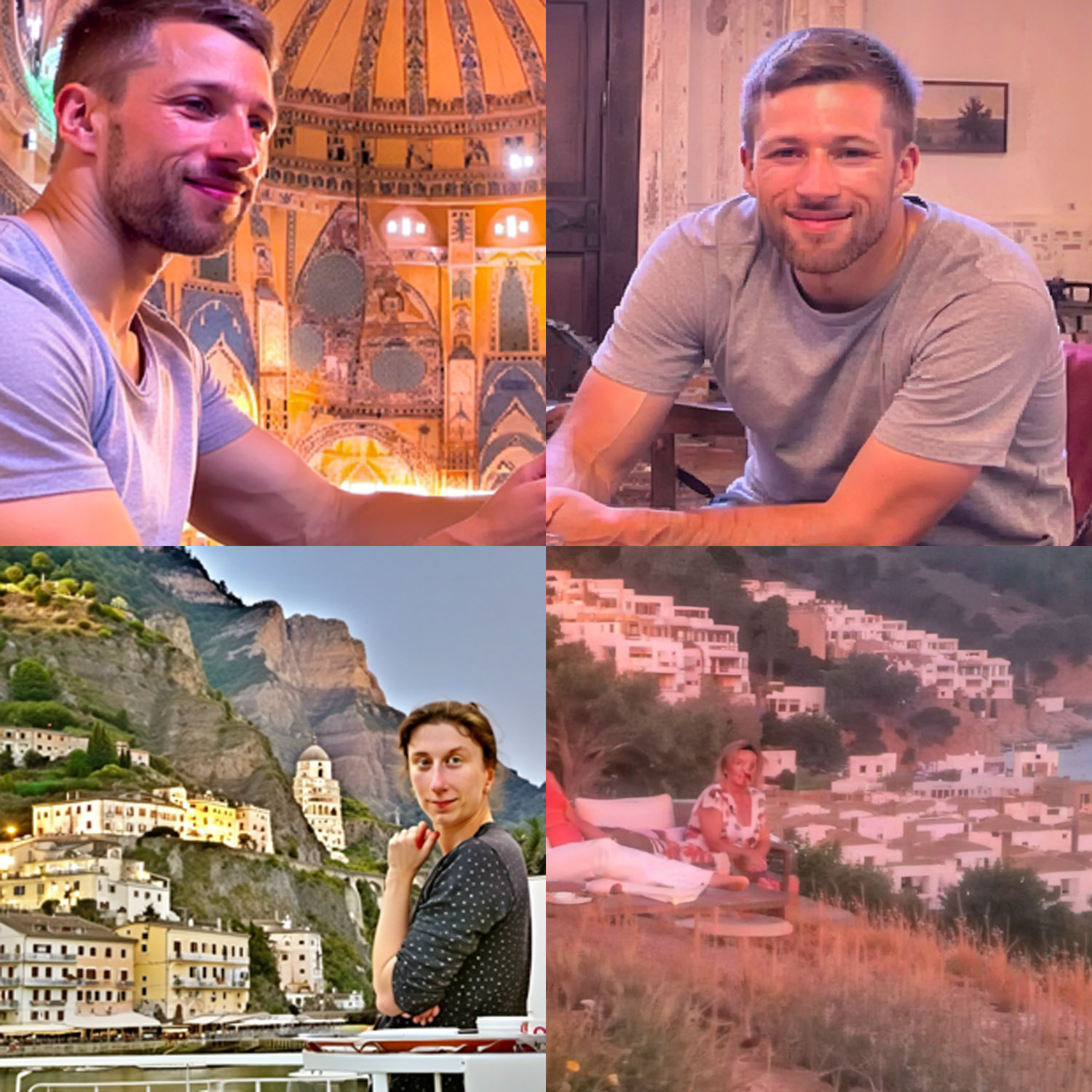}
    \caption{Artifacts in DreamBooth training by using external image collections as regularization datasets.}
    \label{fig:dreambooth-regularization}
\end{figure}

The consequences of this prior destruction were severe: training either diverged, resulting in heavily deformed objects, or the model severely overfitted the subject dataset. Figure~\ref{fig:dreambooth-regularization} illustrates the model’s behavior prior to the breakdown of the class prior. A distinct qualitative indicator of a destroyed class prior was the appearance of a purple tint in generated class-word images (e.g., ``a $[V]$ man'').

This negative result provides strong empirical support for DreamBooth’s autogenous prior preservation loss \cite{Ruiz2022Aug}. Our findings highlight that drastically altering the class prior using external regularization datasets is generally infeasible. Instead, finetuning the model remains the preferred method to shift the class prior. This case study thus serves as a qualitative ablation, illustrating the pitfalls of alternative regularization strategies and reaffirming DreamBooth’s PPL \cite{Ruiz2022Aug}.

\end{document}